\documentclass[runningheads]{llncs}

\usepackage{eccv}

\usepackage{eccvabbrv}
\usepackage{graphicx}
\usepackage{booktabs}
\usepackage{multirow} 
\usepackage{wrapfig}
\usepackage{pifont}
\newcommand{\xmark}{\ding{55}}%
\usepackage[accsupp]{axessibility}  

\usepackage[breaklinks,colorlinks,citecolor=eccvblue]{hyperref}

\usepackage{orcidlink}
\usepackage[table]{xcolor}
\usepackage{listings}
\usepackage{tcolorbox}
\tcbuselibrary{listings}
\usepackage{xcolor}

\definecolor{commentgreen}{RGB}{34, 139, 34}
\definecolor{keywordblue}{RGB}{0, 0, 200}
\definecolor{stringorange}{RGB}{200, 100, 0}

\lstdefinestyle{mystyle-tsne}{
    basicstyle=\ttfamily\small,
    commentstyle=\color{commentgreen}\itshape,   
    keywordstyle=\color{keywordblue}\bfseries,   
    stringstyle=\color{stringorange},            
    morekeywords={for, in, save, plot},          
    morecomment=[l]{\#},                         
}
\lstdefinestyle{mystyle-ug}{
    basicstyle=\ttfamily\scriptsize,
    commentstyle=\color{commentgreen}\itshape,   
    keywordstyle=\color{keywordblue}\bfseries,   
    stringstyle=\color{stringorange},            
    morekeywords={for, in, save, plot},          
    morecomment=[l]{\#},                         
}

\begin{document}



\title{
Semantic Generative Tuning for Unified Multimodal Models
}

\titlerunning{Semantic Generative Tuning for Unified Multimodal Models}

\author{
Songsong Yu\inst{1,2} \and
Yuxin Chen\inst{2} \and
Ying Shan\inst{2} \and
Yanwei Li\inst{1}\textsuperscript{\dag}
}
\authorrunning{Songsong Yu et al.}

\institute{
\textsuperscript{1}Shanghai Jiao Tong University      \textsuperscript{2}Tencent ARCLab
}

\maketitle
\begingroup
\renewcommand{\thefootnote}{\dag}
\footnotetext{Corresponding author.
}
\endgroup

\begin{abstract}
Unified multimodal models (UMMs) strive to consolidate visual understanding and visual generation within a single architecture. However, prevailing training paradigms independently optimize understanding via sparse text signals and generation through dense pixel objectives. Such a decoupled strategy yields misaligned representation spaces, isolating visual understanding from generation and hindering their mutual reinforcement. This work presents the first systematic investigation into generative post-training, where we formulate hierarchical visual tasks as generative proxies to bridge the isolation in UMMs. Our empirical investigation reveals that high-level semantic tasks, particularly image segmentation, serve as optimal proxies. Unlike low-level tasks that distract models with texture details, segmentation provides structural semantics that significantly enhance both vision-centric perception and generative layout fidelity. Building upon these insights, we introduce \textbf{S}emantic \textbf{G}enerative \textbf{T}uning (\textbf{SGT}), a novel paradigm that leverages segmentation as a generative proxy to align and synergize multimodal capabilities. Mechanistic analyses further demonstrate that SGT fundamentally improves feature linear separability and optimizes visual-textual attention allocation pattern. Extensive evaluations show that SGT consistently improves both multimodal comprehension and generative fidelity across mainstream benchmarks. Our code is available on the \href{https://song2yu.github.io/SGT/}{Project Page}.

   \and Generative Tuning
\end{abstract}
\section{Introduction}
\begin{figure*}[t]
    \centering
    \includegraphics[width=\linewidth]{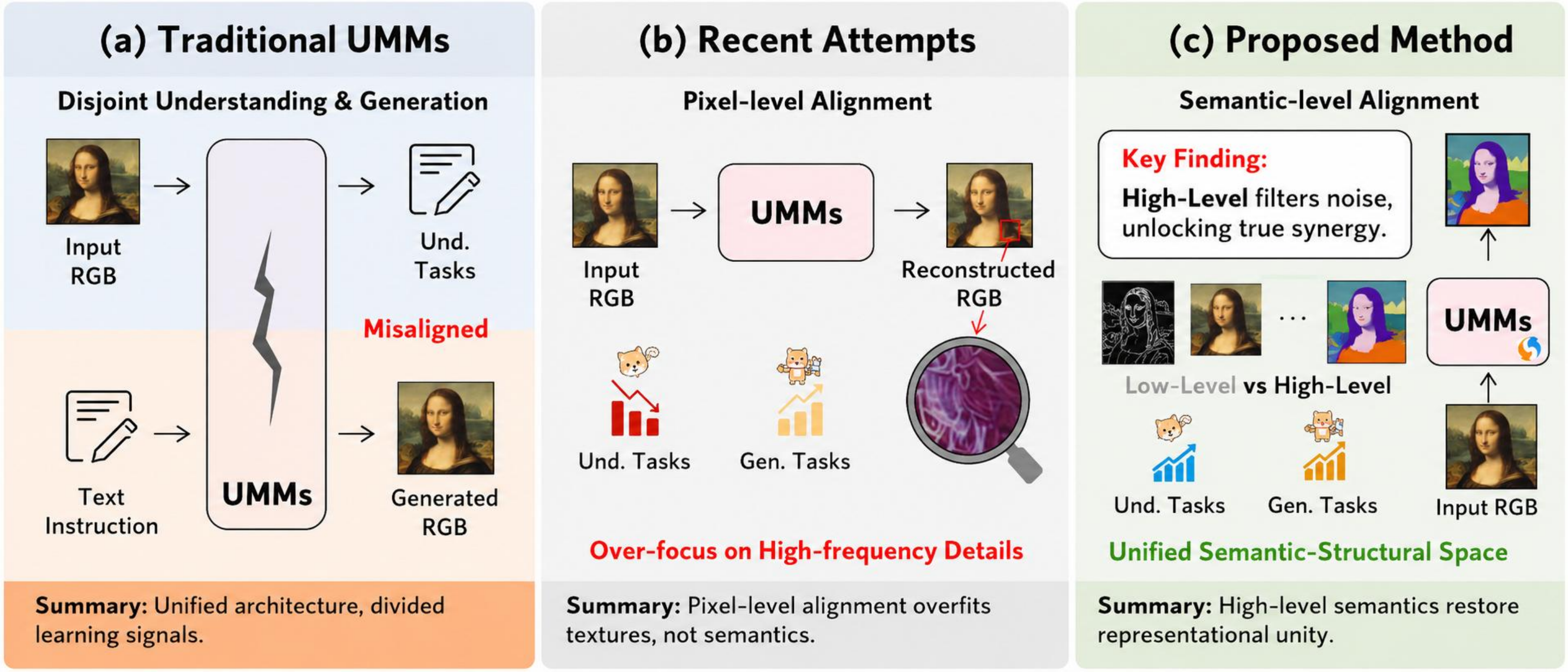}
    \caption{\textbf{Comparison of alignment strategies for UMMs.} 
    (a) Traditional UMMs optimize understanding and generation tasks separately, resulting in low synergy. 
    (b) Recent pixel-level attempts~\cite{dis:reca} over-focus on high-frequency details, bringing suboptimal alignment. 
    (c) Our proposed SGT achieves semantic-level alignment, filtering low-level noise and enabling true synergy between understanding and generation.}
    \label{fig:teaser}
\end{figure*}

The rapid progress of multimodal models~\cite{sora,llava,Infinity,VAR,ju2025video2bev} has been fundamentally shaped by distinct research trajectories for understanding and generation. 
For understanding, models like LLaVA\cite{llava} formulate visual comprehension as a text-generation process, leveraging cross-modal alignment to map visual features into linguistic spaces for complex understanding and reasoning.
As for generation, studies emphasize generative modeling~\cite{sdv3,sora,yu2025mono2stereo}, where diffusion-based architectures have established state-of-the-art performance in high-fidelity content synthesis. While these specialized architectures exhibit significant proficiency within their respective domains, the emergent trend toward UMMs seeks to consolidate both visual comprehension and generation within a single streamlined framework~\cite{umms:li2025uniworldv2,umms:MetaQueries,umms:janus,umms:wang2025skywork,umms:yang2025mmar}. This architectural convergence holds the potential to facilitate the transfer of bidirectional knowledge and foster mutual reinforcement between understanding and generation~\cite{umms:dreamllm,umms:instructblip,umms:janusflow,umms:jin2024unified,umms:jin2024video}. Consequently, this deep integration unlocks advanced capabilities, including interleaved image-text generation and in-context visual editing, establishing a robust foundation for general-purpose multimodal systems~\cite{umms:lmfusion,umms:wise}.

Despite the structural unification, prevailing training paradigms optimize understanding and generation through divergent supervisory signals as shown in Fig.~\ref{fig:teaser}(a). Understanding tasks are predominantly driven by sparse text supervision (e.g., VQA datasets), while generative capabilities are optimized via low-level visual objectives (e.g., pixel or visual token reconstruction). This decoupled training strategy isolates two capabilities and hinders the model from capturing the inherent dependencies between visual understanding and generation. Consequently, UMMs often fail to achieve true mutual reinforcement, leaving the framework with a shared architecture but disjointed optimization processes.

As illustrated in Fig.~\ref{fig:teaser}(b), recent attempts~\cite{dis:reca} address this optimization divergence by employing visual reconstruction in the pixel space as a proxy task. Although this approach yields measurable improvements in generative capabilities, it remains questionable whether low-level visual reconstruction serves as the optimal proxy for synergizing understanding and generation. Since robust visual comprehension inherently relies on semantic information rather than the memorization of low-level textures~\cite{ijepa}, optimizing for pixel-perfect reconstruction compels the architecture to focus on irrelevant granular details.
This distraction inherently limits the model's capacity to enhance visual understanding.

To resolve this critical inquiry, we conduct the first systematic investigation to evaluate the efficacy of various visual proxies in coupling understanding and generation as shown in Fig.~\ref{fig:gain_understanding} and Fig.~\ref{fig:gain_generation}. Specifically, we establish a hierarchical taxonomy of visual objectives comprising low-level, mid-level, and high-level tasks. Each level encapsulates distinct degrees of spatial granularity and semantic information. This empirical investigation reveals that high-level semantic tasks, particularly image segmentation, serve as the optimal proxy. Unlike low-level tasks that over-emphasize textures, segmentation inherently aligns with the semantic demands of visual comprehension.

Guided by these findings, we introduce \textbf{S}emantic \textbf{G}enerative \textbf{T}uning (\textbf{SGT}) for UMMs, as illustrated in Fig.~\ref{fig:teaser}(c). 
This training paradigm leverages image segmentation as a generative proxy to tightly couple visual understanding and generation. To elucidate the underlying mechanisms, we investigate feature distributions and attention dynamics. Our analysis reveals that SGT fundamentally improves feature linear separability and optimizes visual-textual attention allocation. Consequently, this framework effectively enhances both vision-centric perception and generative layout fidelity across mainstream architectures and benchmarks.

The main contributions of this work are summarized as follows.
\begin{itemize}
\item We systematically explore generative tuning by formulating various visual tasks as generative proxies. Our analysis reveals that high-level semantic tasks, particularly image segmentation, significantly outperform low-level reconstruction in synergizing visual understanding and generation.

\item Guided by these insights, we introduce SGT, a novel paradigm that leverages segmentation as a generative proxy to synergize multimodal capabilities. Mechanistic analyses further demonstrate that SGT fundamentally improves feature separability and optimizes visual-textual attention allocation.

\item Extensive evaluations across mainstream UMM architectures validate the efficacy of SGT. By effectively mitigating representational misalignment, the proposed paradigm yields consistent improvements in both visual understanding and generation across diverse benchmarks. Specifically, the framework achieves a 6.02\% performance increase over BAGEL~\cite{bagel} on the CV-Bench~\cite{bench:CV-bench} evaluation and attains a 90.0\% score on the GenEval~\cite{bench:geneval}.

\end{itemize}
\section{Related Work}

\subsection{Unified Multimodal Models} 
Recent UMMs~\cite{umms:liquid,umms:uio2,umms:unitok,umms:vila} focus on any-to-any processing within a single backbone through two primary trajectories. The first trajectory~\cite{umms:seedx,umms:emu3} utilizes discrete visual tokenization and decoder-only autoregression to implement a unified next-token prediction framework. Models such as Emu3~\cite{umms:emu3}, Janus-Pro~\cite{umms:januspro}, and VARGPT~\cite{umms:zhuang2025vargpt} support interleaved reasoning and mixed-modal generation through this paradigm. The second trajectory~\cite{omnigen2, lightbagel, bagel} employs hybrid architectures that combine causal language modeling with denoising objectives to maintain synthesis quality while unifying reasoning, as demonstrated by Show-o~\cite{umms:showo,umms:showo2} and Transfusion~\cite{umms:transfusion}. Research on representation and fusion, including TokenFlow~\cite{umms:qu2025tokenflow} and Chameleon~\cite{umms:chameleon}, further addresses the balance between semantic abstraction and structural integrity. These works collectively demonstrate that unified training and architectural convergence are essential for bridging the gap between semantic understanding and high-fidelity generation.

\subsection{Representation Learning via Generative Objectives}
Recent research has explored the utility of generative models, particularly diffusion~\cite{sdv3,parihar2024precisecontrol,weng2024fast,fu2024geowizard}, for visual representation learning~\cite{vqrae-wangxg,REG-ming,repa-xie}. Initial approaches~\cite{augmentation:luo2024deem,augmentation:shipard2023diversity,augmentation:tian2023stablerep} utilize diffusion models as data augmenters to synthesize diverse training samples, thereby improving zero-shot classification and downstream recognition performance. Beyond data augmentation, several frameworks~\cite{self_supervised:chen2024deconstructing,self_supervised:fuest2024diffusion,self_supervised:graikos2024learned,self_supervised:hudson2024soda,self_supervised:wei2023diffusion} reformulate generative processes as self-supervised objectives. For instance, SODA~\cite{self_supervised:hudson2024soda} optimizes semantic features through a diffusion-based bottleneck, while DDAE~\cite{self_supervised:wei2023diffusion} interprets diffusion as a form of masked autoencoding for reconstruction-based learning. Recent evidence~\cite{semantic:yang2023diffusion,semantic:wang2023infodiffusion,semantic:zhao2023unleashing,li2024selm,wang2025spatialviz} further indicates that intermediate generative features capture rich semantic information that can complement contrastive representations or be directly transferred to recognition tasks. While existing efforts primarily focus on pixel-space reconstruction~\cite{dis:reca,dis:ross,dis:genhancer,yu2024dme,ju2026instruction} to bolster visual representations for recognition or synthesis, our work introduces a systematic investigation into how classical visual tasks influence UMMs.

\subsection{Reconstruction for Understanding and Alignment}

Existing frameworks such as ReCA \cite{dis:reca}, DIVA \cite{dis:diva}, ROSS \cite{dis:ross}, and GenHancer \cite{dis:genhancer} rely on exact pixel reconstruction to enhance model performance. We fundamentally diverge from this paradigm by abandoning raw pixel recovery to eliminate inherent representational redundancy. Crucially, we present the first systematic validation of how hierarchical visual proxy tasks impact the generative tuning of UMMs. By establishing this comprehensive taxonomy, we conclusively demonstrate that advanced visual tasks deliver the maximum performance improvements. Furthermore, while contemporary studies like UniMRG \cite{dis:UniMRG} explore isolated proxy tasks and Metamorph \cite{dis:metamorph} observes the mutual influence between perception and synthesis, our work actively bridges the gap between discriminative and generative capabilities. This unified optimization explicitly establishes a shared semantic space to capture the structural abstraction essential for general purpose multimodal learning.

\section{Semantic Generative Tuning}
This section outlines the whole framework. It begins by formalizing the preliminaries of UMMs in Sec.~\ref{sec:medthod_prelims}. Then, Sec.~\ref{sec:medthod_architecture} details the training strategies applied to representative architectures such as BAGEL~\cite{bagel} and OmniGen2~\cite{omnigen2}. For systematic evaluation over understanding and generative capabilities, Sec.~\ref{sec:medthod_3findings} introduces a hierarchical suite of tasks within a generative tuning framework and assesses their influence on six core understanding metrics as well as generative performance.

\subsection{Formulation}
\label{sec:medthod_prelims}
UMMs aim to integrate diverse modalities within a single architecture $f_{\theta}$ by mapping inputs from the textual space $\mathcal{T}$ and image space $\mathcal{I}$ into a shared representation space. Formally, given a text prompt $x \in \mathcal{T}$ and an optional reference image $v \in \mathcal{I}$, the model processes various tasks through different input combinations. For visual understanding tasks, UMMs typically process an input image using a semantic vision encoder and subsequently integrate the extracted features with language tokens for unified treatment within a language model. In the case of visual editing tasks, certain frameworks~\cite{bagel,omnigen2,umms:januspro,umms:MetaQueries,umms:openuni} supplement the semantic vision encoder with a variational autoencoder (VAE) to preserve fine-grained image details as well as to ensure identity consistency and high-quality generation.

Without loss of generality, we employ a dual encoder architecture as an illustrative example to introduce the general formulation of UMMs. Specifically, a ViT-based encoder $\Phi_{vit}(\cdot)$ extracts semantic tokens $z_{vit} \in \mathbb{R}^{L \times D}$ for multimodal reasoning, while a VAE-based encoder $\Phi_{vae}(\cdot)$ encodes the image into a latent space $z_{vae} \in \mathbb{R}^{H \times W \times C}$ to maintain structural and textural details. The mapping for these tasks is formulated as follows

\begin{equation}
    y = 
    \begin{cases} 
    f_{\theta}(x, [z_{vit}]) & \text{Understanding: } y \in \mathcal{T} \\
    f_{\theta}(x, [z_{noise}]) & \text{Generation: } y \in \mathcal{I} \\
    f_{\theta}(x, [z_{vit}, z_{vae}, z_{noise}]) & \text{Editing: } y \in \mathcal{I}
    \end{cases}
\label{eq:umms}
\end{equation}

where $[\cdot]$ denotes the set of optional inputs and $z_{noise}$ represents the initial Gaussian noise utilized for generative processes. This formulation categorizes the operational scope of UMMs into three distinct functional paradigms. For visual understanding, the model leverages semantic features $z_{vit}$ to generate textual responses $y \in \mathcal{T}$. In the context of visual generation, the model maps a text prompt $x$ and the initial noise $z_{noise}$ to a synthesized image $y \in \mathcal{I}$. For visual editing tasks, the framework integrates $z_{vit}$, $z_{vae}$, and the stochastic component $z_{noise}$ to achieve high-fidelity image manipulation. Such a structure simultaneously yields representations across varying granularities to establish a robust foundation for UMMs.

\subsection{Motivation and Hierarchical Visual Task Taxonomy}
\label{sec:medthod_architecture}
Recent advances~\cite{dis:ross,dis:genhancer,umms:unihetero,vapi2025,dis:diva} indicate that reconstructing visual inputs from learned embeddings significantly enhances the representation quality of visual embeddings. However, pixel-space reconstruction fundamentally optimizes image fidelity rather than cross-modal semantic alignment, and its objective is not invariably the most relevant for visual understanding and reasoning. Driven by this insight, we pose the question of whether pixel-space reconstruction is truly the optimal choice for UMMs.

\begin{figure}[t]
  \centering
  \includegraphics[width=0.96\linewidth]{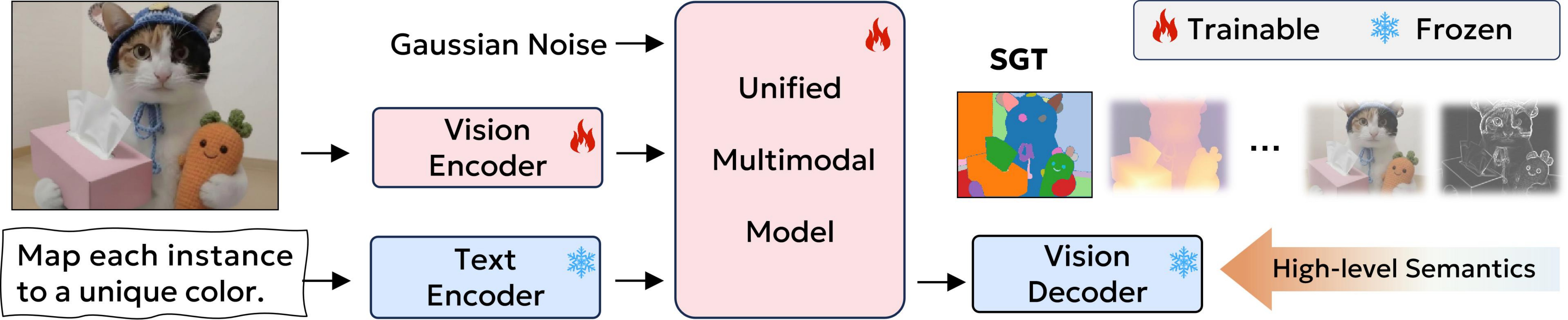}
  \caption{
    Overview of the generative tuning paradigm. An RGB image and a concise textual instruction are processed by respective vision and text encoders to extract independent embeddings. UMMs then integrate these embeddings and map the representations to the designated task. Because empirical evaluations demonstrate that visual generation targets at an advanced semantic level yield the most significant performance gains, SGT explicitly adopts image segmentation as its generative objective.
  }
  \label{fig:unified_mm}
\end{figure}

In response to this question, we establish a hierarchical taxonomy to investigate the impact of different levels of visual tasks on UMMs within the generative tuning framework. Formally, we model the generative tuning as a conditional generation process $y = f_{\theta}(x,\allowbreak [z_{\text{vit}}, z_{\text{noise}}])$,  where the output $y$ resides in the visual space. We define the training objective as $\mathcal{L} = L(f_{\theta}(x, [z_{vit}, z_{noise}]), \hat{y})$, where $x$ denotes a concise natural language instruction tailored to the specific task, and $\hat{y}$ represents the target visual representation as depicted in Fig.~\ref{fig:unified_mm}. Here, $\hat{y}$ denotes the ground truth for diverse visual tasks. Crucially, to isolate the impact of task granularity, we exclusively utilize visual data for generative tuning during this investigative phase, strictly excluding other data types such as visual question answering, text-to-image generation, or standard image editing data. To ensure a rigorous comparison, all tasks are evaluated using the same set of input RGB images and an identical volume of training data. Specifically, our evaluation covers high-level tasks (segmentation, object detection), mid-level tasks (depth estimation, inpainting), and low-level tasks (edge detection). Detailed data processing procedures are provided in the supplementary material.

\subsection{From Empirical Observations to the SGT Paradigm
}
\label{sec:medthod_3findings}
\begin{figure}[t] 
\centering
\begin{subfigure}{0.95\linewidth}
    \centering
    \includegraphics[width=\linewidth]{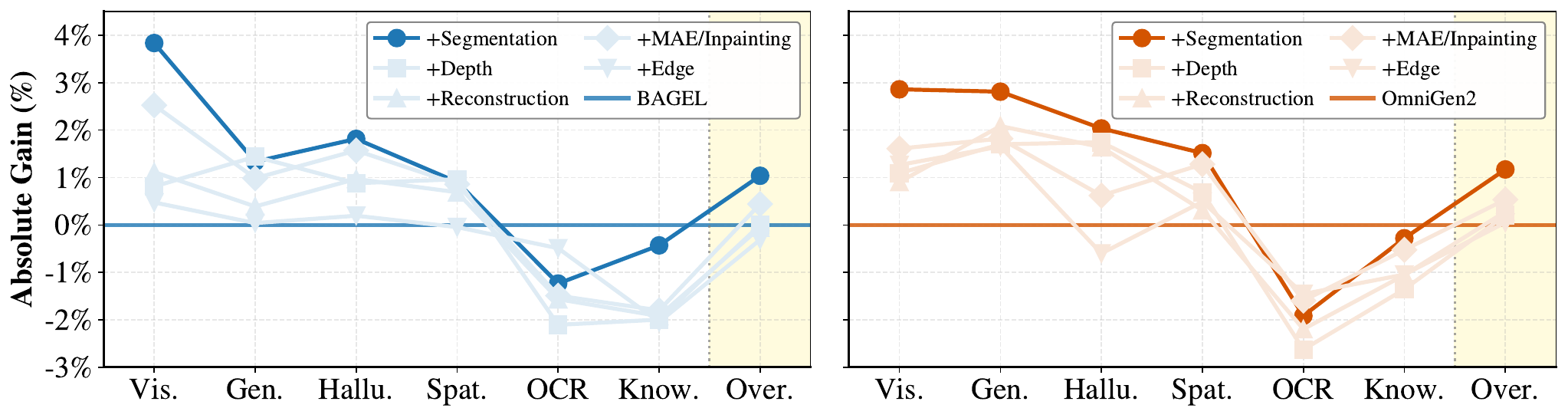}
    \caption{Understanding capability gains}
    \label{fig:gain_understanding}
\end{subfigure}

\begin{subfigure}{0.95\linewidth}
    \centering
    \includegraphics[width=\linewidth]{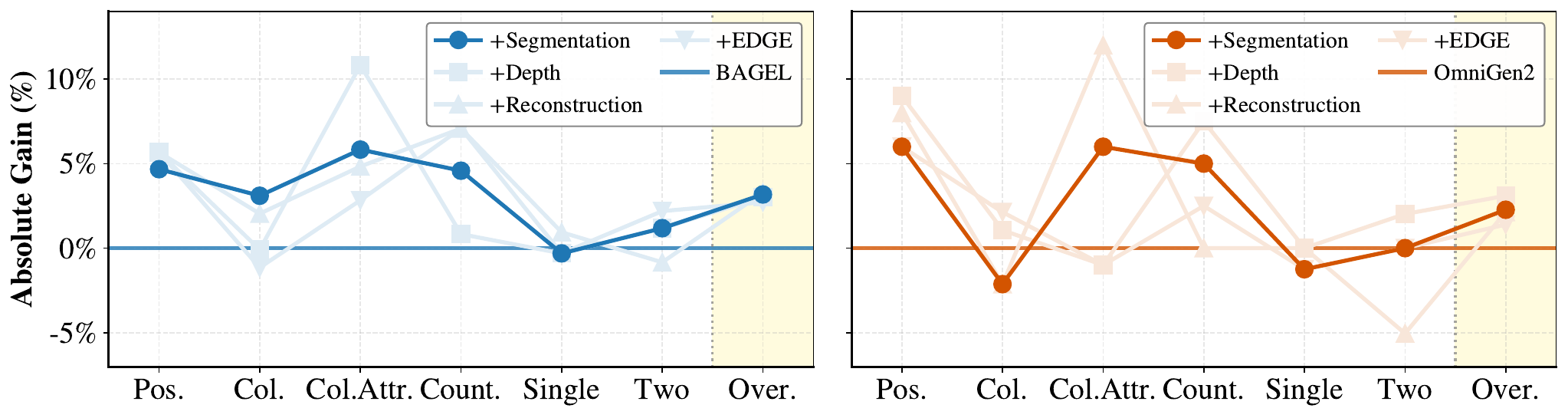}
    \caption{Generation capability gains}
    \label{fig:gain_generation}
\end{subfigure}

\caption{Empirical evaluation of the hierarchical task ladder across diverse understanding and generation dimensions. 
(a) High-level proxy tasks yield greater performance gains than low-level tasks in multimodal understanding. 
(b) Various generative objectives consistently improve performance in the position dimension, yielding comparable overall gains. (From left to right): Position, Colors, Color Attributes, Counting, Single Object, Two Objects, and Overall. The results represent the average performance computed across twelve random seeds.
}
\label{fig:hierarchical_comparison}
\end{figure}

We begin by evaluating visual proxy tasks across different levels based on empirical model performance variations. To establish a comprehensive and systematic evaluation protocol, we draw inspiration from the taxonomy proposed in Cambrian-1~\cite{bench:CV-bench}. Specifically, we augment the original categories of general VQA~\cite{bench:mmmu,bench:mmstar}, vision-centric perception~\cite{bench:CV-bench,bench:MMVP}, chart/OCR~\cite{bench:ocrbench,bench:docvqa}, and mathematical reasoning~\cite{bench:mathvista,bench:scienceqa} with spatial reasoning~\cite{bench:VSR,bench:sibench} and hallucination resistance~\cite{bench:pope,bench:hallusion} to enable a more holistic assessment. Each capability score is derived from the unweighted average of two representative benchmarks. Generative capabilities are evaluated via GenEval~\cite{bench:geneval}. We validate our findings across both BAGEL~\cite{bagel} and OmniGen2~\cite{omnigen2} to ensure architectural generalizability, with specific model details provided in Sec.~\ref{sec:exp_setup}. Our empirical analysis yields three crucial observations, as visualized in Fig.~\ref{fig:gain_understanding} and Fig.~\ref{fig:gain_generation}.

\vspace{2.5pt}
\noindent\textbf{Observation 1: High-level semantic tasks outperform low-level cues.} Our analysis indicates that high-level tasks yield substantially greater benefits for multimodal understanding than their mid- or low-level counterparts. As evidenced in Fig.~\ref{fig:gain_understanding}, high-level objectives such as image segmentation consistently outperform mid-level tasks (e.g., depth estimation) and low-level tasks (e.g., edge detection). We attribute this to the strong alignment between high-level semantic and the reasoning requirements of understanding models. High-level supervision encourages the extraction of semantic and structural essence, whereas low-level tasks may compel the model to overfit to intricate textural details that are often redundant for complex reasoning. This observation aligns with findings in GenHancer~\cite{dis:genhancer} and the design philosophy of I-JEPA~\cite{ijepa}.

\vspace{2.5pt}
\noindent\textbf{Observation 2: Visual supervision enhances perception, not reasoning.} The generative tuning paradigm predominantly fortifies fundamental visual perception rather than linguistic priors or abstract logical reasoning. While we observe significant performance gains in vision-centric tasks, spatial reasoning, and hallucination resistance, capabilities in chart recognition and mathematical knowledge remain static or exhibit marginal decline, as shown in Fig.~\ref{fig:gain_understanding}. This divergence indicates that while visually-derived supervision enhances representation quality to boost perceptual capabilities, it does not impart additional knowledge or logical reasoning skills.

\vspace{2.5pt}
\noindent\textbf{Observation 3: Various proxy tasks consistently improve spatial fidelity.} Diverging from the trends associated with varying granularities observed in understanding benchmarks, the generative tuning paradigm consistently enhances overall generation quality. Otherwise, as illustrated in Fig.~\ref{fig:gain_generation}, the model demonstrates consistent performance gains on position-aware tasks. This suggests that visual proxy tasks inherently provide explicit spatial constraints, regardless of their semantic granularity. Empirically, the process of reconstructing these visual structures forces the model to maintain accurate spatial layouts, thereby naturally enhancing its alignment with positional prompts. This observation aligns with insights reported in RecA~\cite{dis:reca}.

Synthesizing these three observations, we conclude that within the generative tuning framework, employing high-level semantic proxy tasks for generative tuning yields optimal enhancements for UMMs. Consequently, we advocate for a novel training paradigm termed Semantic Generative Tuning (SGT). This approach strategically leverages high-level visual proxies, especially image segmentation, to refine the internal representations of UMMs, thereby harmonizing visual understanding and generation within a unified framework. Additional experiments show that semantic instance and panoptic segmentation, as well as class-agnostic segmentation, consistently yield comparable improvements. Detailed results are provided in the supplementary materials.

\section{Experiments}

\begin{table}[t]
    \centering
    \caption{Statistics of the training data.}
    \label{tab:data_card}
    \setlength{\tabcolsep}{2mm} 
    \resizebox{\linewidth}{!}{ 
    \begin{tabular}{lcccccc}
        \toprule
        \textbf{Data Source} & SGT & General & Doc/Chart/Screen & Math/Reasoning & General OCR & Language \\
        \midrule
        \textbf{Number}   & 190k & 180k    & 103k             & 101k           & 45k         & 72k      \\
        \bottomrule
    \end{tabular}
    }
\end{table}

We first detail the experimental configurations and the selection of models in Sec.~\ref{sec:exp_setup}. Sec.~\ref{sec:exp_comparison} presents a unified study that (i) benchmarks our approach against state-of-the-art UMMs on diverse understanding and generation tasks and (ii) evaluates alternative visual proxy tasks. Furthermore, we investigate the optimal data recipe and the scaling properties in Sec.~\ref{sec:exp_properties}. In Sec.~\ref{sec:exp_analysis}, we analyze how the SGT paradigm alters the feature space and attention allocation of UMMs, in order to uncover deeper underlying causes.

\subsection{Experimental Setup}
\label{sec:exp_setup}
\textbf{Datasets.} Although Sec.\ref{sec:medthod_3findings} confirms that semantic generative tuning is highly effective in isolation, we further construct a holistic post-training to fully unleash the potential of SGT. By synergizing SGT with 500k supervised fine-tuning samples from LLaVA-OneVision\cite{llava-ov}, we demonstrate its robustness and scalability. To strictly preclude data overlap between the training and evaluation phases, we source all images for SGT exclusively from the SAM~\cite{sam} dataset. Specifically, we curate 190k samples for the SGT dataset, with the detailed source distribution outlined in Table~\ref{tab:data_card}. Regarding the VQA data, we align data mixture with the official recipe provided by LLaVA-OneVision\cite{llava-ov}.

\vspace{2.5pt}
\noindent\textbf{Model selection.} We conduct our experiments on two mainstream UMM architectures, BAGEL~\cite{bagel} and OmniGen2~\cite{omnigen2}, to evaluate our method across distinct design philosophies. Beyond an approximate twofold difference in parameter scale, these models differ fundamentally in their feature interaction mechanisms and training paradigms. Specifically, BAGEL adopts a Mixture of Transformers framework to facilitate layer-wise feature sharing throughout the network. Conversely, OmniGen2 utilizes hidden states from the understanding module as semantic guidance to steer the generative process. Their training strategies also diverge considerably, as BAGEL employs a native interleaved training process, whereas OmniGen2 pairs a frozen pre-trained vision language model~\cite{Qwen2.5-VL} with a diffusion module trained from scratch. To further validate the universality of SGT paradigm beyond these UMMs, we extend our preliminary evaluation to single visual encoder architectures, detailing the results in Supplementary Material. This architectural diversity ensures the broad applicability of SGT.

\vspace{2.5pt}
\noindent\textbf{Evaluation benchmarks.} To comprehensively assess multimodal understanding, we utilize the VLMEvalKit~\cite{vlmevalkit} to evaluate model performance across a diverse suite of benchmarks. This carefully curated selection encompasses spatial reasoning, robustness against hallucinations, general visual question answering, knowledge reasoning, and vision-centric perception to ensure a holistic evaluation. Specifically, we conduct these assessments on CV-Bench~\cite{bench:CV-bench}, MMVP~\cite{bench:MMVP}, VSR~\cite{bench:VSR}, SIBench-mini~\cite{bench:sibench}, POPE~\cite{bench:pope}, Hallusion~\cite{bench:hallusion}, MMBench-TEST-EN 1.1~\cite{bench:mmbench}, MMMU-val~\cite{bench:mmmu}, RWQA~\cite{bench:xai2024grok15v}, MathVista~\cite{bench:mathvista}, BLINK~\cite{bench:blink}, MME~\cite{bench:mme}, and MMStar~\cite{bench:mmstar}. Furthermore, we employ GenEval~\cite{bench:geneval} and GEdit-Bench-En~\cite{bench:gedit} to measure text-to-image generation and image editing capabilities respectively. We detail the optimization process and hardware configurations in the supplementary material.

\subsection{Main Results}
\label{sec:exp_comparison}

\begin{table*}[t]
\centering
\small
\setlength{\tabcolsep}{3pt}
\caption{{Comparison with state-of-the-art UMMs.} Best results are in \textbf{bold}, second best are \underline{underlined}. ``--'' indicates not reported. \xmark~indicates the model does not support image editing. \dag~refers to methods using LLM rewriter. $^*$Results are taken from previous works. $\ddag$ indicates that the first term denotes understanding parameters, and the second denotes generation parameters.
}
\label{tab:sota_comparison}
\vspace{-2mm}
\resizebox{\textwidth}{!}{%
\begin{tabular}{@{}lc|cccccc|cc@{}}
\toprule
\multirow{2}{*}{\textbf{Model}} & \multirow{2}{*}{\textbf{Params}} & \multicolumn{6}{c|}{\textbf{Visual Understanding}} & \multicolumn{2}{c}{\textbf{Visual Generation}} \\
 &  & MMVP & VSR & Hallu. & MMStar & RWQA & MathV. & GenEval & GEdit-Bench-En \\
\midrule
\multicolumn{10}{c}{\small \em \textit{Small-scale Models ($\leq$4B)}} \\
\midrule
$^*$Show-o$_{512}$~\cite{umms:showo} & 1.3B & 50.00 & 54.26 & 46.06 & -- & 38.17 & -- & 68.0 & \xmark \\
Harmon~\cite{umms:harmon} & 1.5B & 60.00 & 60.88 & 46.69 & 38.00 & 48.00 & 33.70 & 73.0 & \xmark \\
ReCA-Harmon~\cite{dis:reca} & 1.5B & 47.00 & -- & 36.70 & 25.53 & 43.53 & 24.50 & \textbf{90.0} & \xmark \\
$^*$UniLIP~\cite{umms:unilip} & 2B & \underline{73.00} & 65.55 & 60.57 & -- & 64.18 & -- & \textbf{90.0} & -- \\
$^*$UniMRG~\cite{dis:UniMRG} & 3.6B & \textbf{74.67} & 73.90 & \textbf{64.56} & -- & \textbf{66.01} & -- & 55.8 & \xmark \\
$^*$OpenUni~\cite{umms:openuni} & 2B & {71.67} & 66.69 & 60.88 & -- & \underline{65.23} & -- & 51.0 & - \\
OmniGen2~\cite{omnigen2} & 3B+4B$\ddagger$ & 65.00 & \underline{77.52} & 62.35 & \underline{55.07} & 64.41 & \underline{63.50} & {76.6} & \underline{6.63} \\
\midrule
\rowcolor{gray!20}SGT-Gen2 & 3B+4B$\ddagger$ & 68.33 & \textbf{78.85} & \underline{64.25} & \textbf{57.07} & 65.10 & \textbf{64.00} & \underline{78.9} & \textbf{6.83} \\
\midrule
\multicolumn{10}{c}{\small \em \textit{Large-scale Models ($\geq$7B)}} \\
\midrule
Chameleon~\cite{umms:chameleon} & 7B & 50.00 & -- & 31.13 & 28.93 & 39.00 & 21.90 & 39.0 & \xmark \\
Janus-Pro~\cite{umms:januspro} & 7B & 63.00 & 71.03 & 60.15 & 46.80 & 41.83 & 42.60 & 80.0 & \xmark \\
$^*$Emu3~\cite{umms:emu3} & 8B & -- & -- & -- & -- & 57.40 & -- & 66.0\dag & \xmark \\
UniWorld-v1~\cite{umms:lin2025uniworld} & 7B+12B$\ddagger$ & 77.67 & \textbf{83.34} & \underline{68.35} & 63.90 & 67.58 & 68.20 & 84.0\dag & 4.85 \\
BAGEL~\cite{bagel} & 7B+7B$\ddagger$ & \underline{83.00} & 80.45 & 68.34 & \underline{67.46} & \underline{71.26} & \underline{73.10} & \underline{88.0\dag} & \underline{6.64} \\
\midrule
\rowcolor{gray!20}SGT-BAGEL & 7B+7B$\ddagger$ & \textbf{83.33} & \underline{81.54} & \textbf{70.24} & \textbf{68.33} & \textbf{72.42} & \textbf{73.90} & \textbf{90.0}\dag & \textbf{6.94} \\
\bottomrule
\end{tabular}%
}
\vspace{-10pt}
\end{table*}

\begin{table*}[t]
    \centering
    \caption{{Unified performance comparison on various benchmarks.} The best results in each group are highlighted in \textbf{bold}. The results reported for the GenEval benchmark represent the average performance computed across twelve random seeds.}
    \label{tab:main_results_ablation}
    \resizebox{\textwidth}{!}{%
        \begin{tabular}{lccccccccccc}
            \toprule
            & \multicolumn{2}{c}{\textbf{Vision-Centric}} & \multicolumn{2}{c}{\textbf{Spatial Reasoning}} & \multicolumn{2}{c}{\textbf{Hallucination}} & \multicolumn{3}{c}{\textbf{General}} & \multicolumn{2}{c}{\textbf{Generation}} \\
            \cmidrule(lr){2-3} \cmidrule(lr){4-5} \cmidrule(lr){6-7} \cmidrule(lr){8-10} \cmidrule(lr){11-12}
            \textbf{Method} & CV-Bench & MMVP & VSR & SIBench & POPE & Hallusion & MMBench & MMMU & MMStar & GenEval & GEdit-Bench-En \\
            \midrule
            \multicolumn{12}{c}{\small \em Base Model: OmniGen2} \\
            \midrule
            OmniGen2 (Base)& 65.94 & 65.00 & 77.52 & 43.29 & 85.97 & 62.35 & 77.04 & 42.11 & 55.07 & 76.58 & 6.63 \\
            SFT            & 65.99 & 66.00 & 77.61 & 44.37 & 86.25 & 64.35 & 77.04 & 43.22 & 55.73 & 74.54 & 6.32 \\
            SFT+Edge       & 66.67 & 65.33 & 77.99 & \textbf{45.51} & 86.10 & 63.72 & 76.88 & 42.78 & 55.53 & {77.45} & 6.79 \\
            SFT+Reconstruction       & 66.71 & 66.33 & 78.18 & 45.41 & 85.92 & \textbf{65.19} & 77.00 & 44.44 & 55.73 & 77.53 & 6.81 \\
            \rowcolor{gray!20}SFT+SGT    & \textbf{66.91} & \textbf{68.33} & \textbf{78.85} & 45.37 & \textbf{87.29} & 64.25 & \textbf{77.09} & \textbf{45.89} & \textbf{57.07} & \textbf{78.86} & \textbf{6.83} \\
            \midrule
            \multicolumn{12}{c}{\small \em Base Model: BAGEL} \\
            \midrule
            BAGEL (Base)   & 73.21 & 83.00 & 80.45 & 48.95 & 85.69 & 68.34 & 81.25 & 46.77 & 67.46 & 78.21 & 6.52 \\
            SFT            & 74.61 & 82.67 & 80.69 & 49.34 & 86.77 & 67.92 & 81.86 & 47.33 & 66.93 & 77.18 & 6.49 \\
            SFT+Edge       & 74.56 & \textbf{83.67} & 80.83 & 49.51 & 86.48 & 68.66 & 81.89 & 47.56 & 67.20 & 79.96 & 6.72 \\
            SFT+Reconstruction       & 75.23 & 83.33 & 80.83 & \textbf{50.59} & 87.98 & 68.03 & 82.18 & 46.56 & 67.40 & 80.82 & {6.75} \\
            \rowcolor{gray!20}SFT+SGT & \textbf{79.23} & 83.33 & \textbf{81.54} & 50.18 & \textbf{88.32} & \textbf{70.24} & \textbf{83.84} & \textbf{48.56} & \textbf{68.33} & \textbf{80.95} & \textbf{6.94} \\
            \bottomrule
        \end{tabular}%
    }
\end{table*}

\begin{figure}[t]
    \centering
    \includegraphics[width=\linewidth]{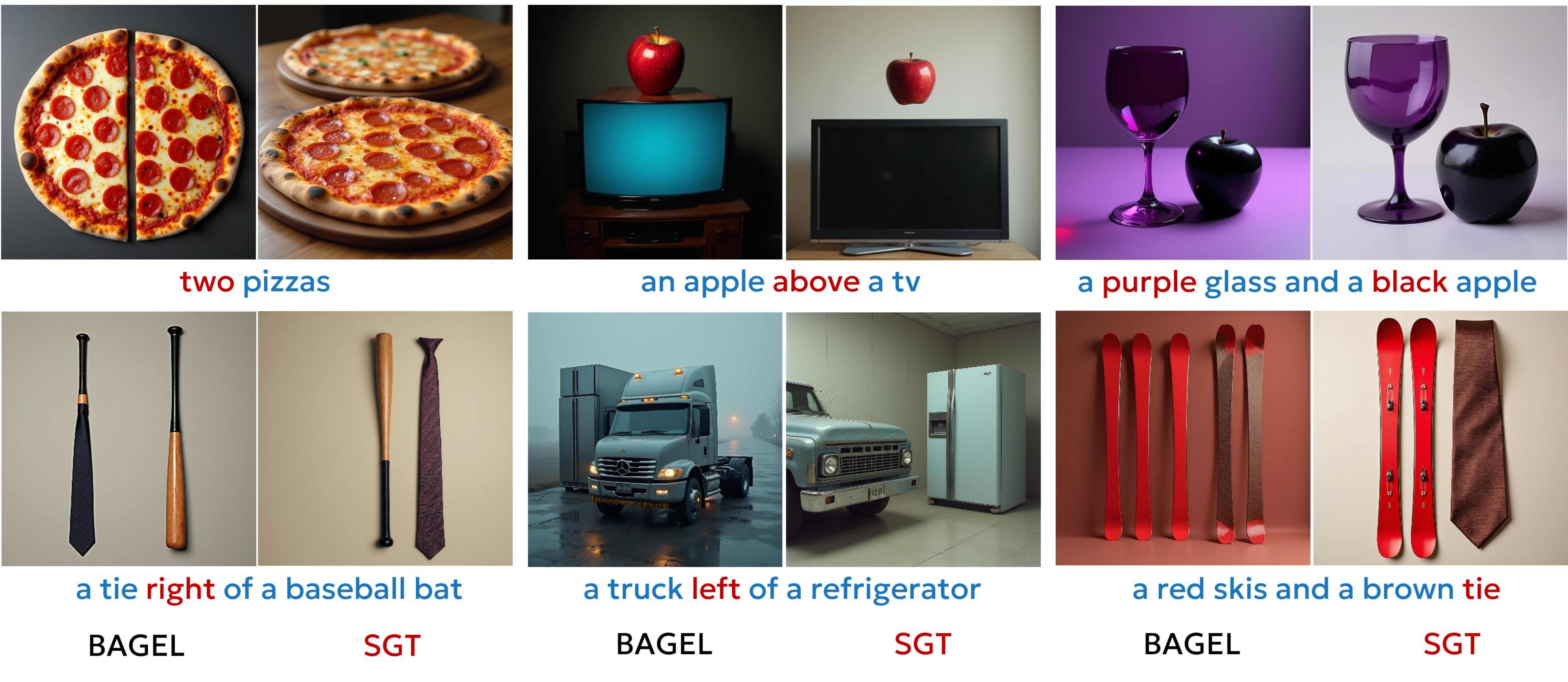}
    \caption{{Qualitative comparison on compositional text-to-image generation.} }
    \label{fig:qualitative_comparison}
\end{figure}

\textbf{Comparison with state-of-the-art UMMs.}
We present a comprehensive comparison between our proposed models and existing leading UMMs in Table~\ref{tab:sota_comparison}. SGT-BAGEL and SGT-Gen2 represent the enhanced variants of BAGEL and OmniGen2. We train these variants using segmentation data from the SAM dataset~\cite{sam} alongside visual understanding instruction tuning. Quantitative evaluations indicate that both SGT-BAGEL and SGT-Gen2 consistently outperform their original baseline architectures and surpass a broad range of competitive models across multiple benchmarks. This widespread superiority demonstrates the efficacy of integrating high-level semantic generative objective into the fine-tuning of UMMs. Furthermore, our framework achieves favorable performance in generative tasks. As Fig.~\ref{fig:qualitative_comparison} illustrates, SGT demonstrates superior adherence to complex textual prompts including spatial and color instructions when compared to the baseline model. Such qualitative improvements confirm that SGT exerts a synergetic benefit on the overall capabilities of UMMs.

\vspace{2.5pt}
\noindent\textbf{Ablation study.} We conduct comprehensive ablation studies to systematically evaluate the isolated impact of SFT data alongside its joint training dynamics with visual tasks across varying semantic levels. As shown in Table~\ref{tab:main_results_ablation}, SFT+SGT utilizes the segmentation task from the SAM dataset as the generative target, whereas SFT+Reconstruction and SFT+Edge employ image reconstruction and edge detection as their respective proxy tasks. All three tasks yield performance gains across the majority of perception-centric understanding benchmarks. We observe notable improvements in vision-focused evaluations such as MMVP and CV-Bench, spatial reasoning assessments including VSR and SIBench-mini, and various hallucination robustness tests. Crucially, SGT yields the most substantial performance gains among the evaluated proxy tasks. This outcome directly corroborates the findings detailed in Sec.~\ref{sec:medthod_3findings} and validates this semantic approach as the optimal target for generative tuning. In generative evaluations, all three proxy tasks achieve comparable gains in text-to-image synthesis, while gains in image editing are positive but smaller in magnitude. This discrepancy suggests that while generative tuning successfully aligns representational spaces, driving further substantial gains in complex generative editing may require the integration of explicit image editing data. Finally, consistent performance improvements observed across both the BAGEL and OmniGen2 architectures underscore the generalizability and robustness of SGT.

\subsection{More Explorations}
\label{sec:exp_properties}

\begin{figure}[t]
    \centering
    \begin{subfigure}[b]{0.48\linewidth}
        \centering
        \includegraphics[width=\linewidth]{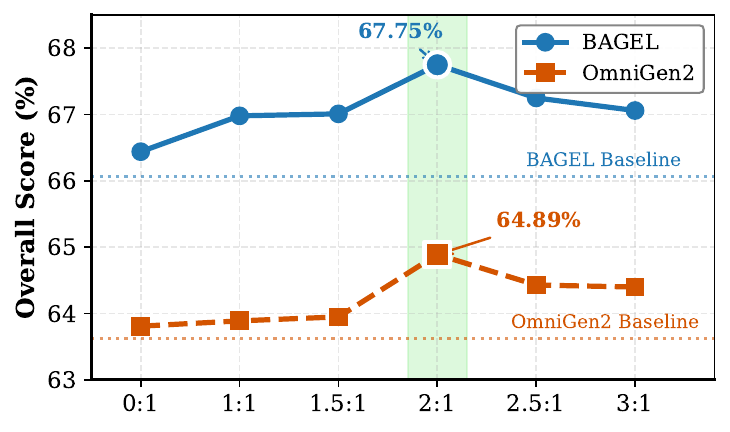}
        \caption{Segmentation-to-VQA Ratio}
        \label{fig:ablation_ratio}
    \end{subfigure}
    \hfill
    \begin{subfigure}[b]{0.48\linewidth}
        \centering
        \includegraphics[width=\linewidth]{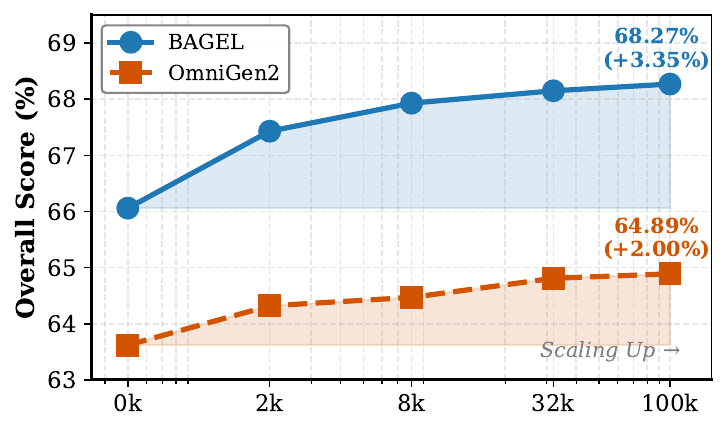}
        \caption{Scalability}
        \label{fig:ablation_scaling}
    \end{subfigure}
    
    \caption{\textbf{Ablation studies on segmentation data integration.} 
    (a) \textbf{The optimal data mixture.} We analyze the Segmentation-to-VQA ratio within each training batch, observing that both models achieve optimal performance at a 2:1 ratio.
    (b) \textbf{Data scalability.} Performance improves consistently as the segmentation dataset expands from 2k to 100k samples (BAGEL: +3.3\%, OmniGen2: +2.0\%), confirming that our visual proxy task yields scalable benefits for multimodal understanding.}
    \label{fig:ablation}
\end{figure}

\noindent\textbf{Optimal data recipe.}
While our analysis in Sec.~\ref{sec:medthod_3findings} indicates that SGT independently enhances both understanding and generation, we posit that a comprehensive post-training regime must synergize SGT objectives with SFT data to maximize performance. Therefore, we conduct an ablation study to determine the optimal data sampling recipe between VQA instructions and segmentation-based visual targets within each training batch. To ensure a robust assessment, we aggregate performance across eight diverse understanding benchmarks~\cite{bench:CV-bench,bench:mmmu,bench:MMVP,bench:mmstar,bench:blink,bench:hallusion,bench:mme,bench:pope} and report the average normalized score. As illustrated in Fig.~\ref{fig:ablation_ratio}, a 1:2 intra-batch ratio of VQA to segmentation data yields the most significant improvements in this aggregate metric for both the BAGEL and OmniGen2 architectures. Regarding generative tasks, we observe that performance scales positively with the proportion of generative samples within the training batch. Balancing these multi-faceted requirements, we adopt the 1:2 ratio for our final configuration. We reserve the exploration of more complex tripartite mixing strategies involving understanding, generation, and SGT for future research.

\begin{figure}[t]
    \centering
    \includegraphics[width=0.99\linewidth]{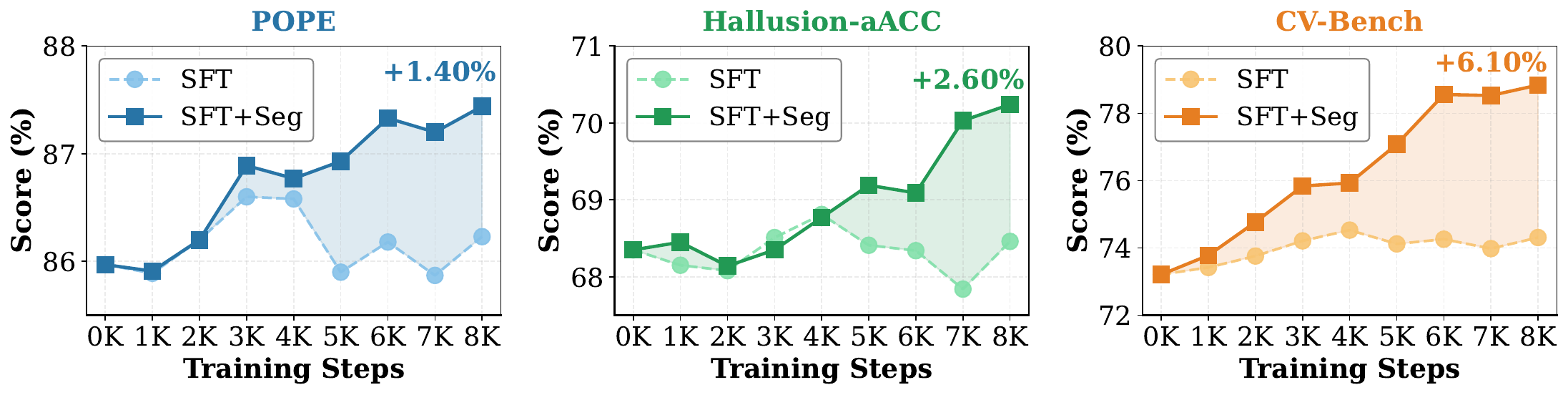}
    \caption{\textbf{Training dynamics with different SFT:Seg ratios.} 
    We compare the training curves of BAGEL under 1:0 (baseline, no segmentation) and 
    1:2 (with segmentation data) ratios across three benchmarks. 
    }
    \label{fig:training_dynamics}
\end{figure}

\vspace{2.5pt}
\noindent\textbf{Scaling properties of SGT.}
To verify the scalability of SGT, we fix the VQA SFT data and systematically scale the segmentation training data. We report the aggregate performance across the eight representative benchmarks described previously. As illustrated in Fig.~\ref{fig:ablation_scaling}, the average normalized score exhibits a monotonic increase commensurate with the volume of segmentation data. Furthermore, an analysis of the training dynamics in Fig.~\ref{fig:training_dynamics} reveals that the integration of segmentation objectives significantly accelerates convergence on challenging benchmarks such as CV-Bench and Hallusion. Compared to the baseline trained exclusively on VQA SFT data, our strategy consistently achieves superior performance during optimization. This demonstrates that SGT serves as a scalable approach to continuously enhance multimodal capabilities.

\subsection{Mechanistic Insights: Why Semantic Proxies Unlock Synergy?}
\label{sec:exp_analysis}
\begin{figure}[t]
    \centering
    \begin{minipage}[c]{0.28\linewidth}
        \centering
        \includegraphics[width=\linewidth]{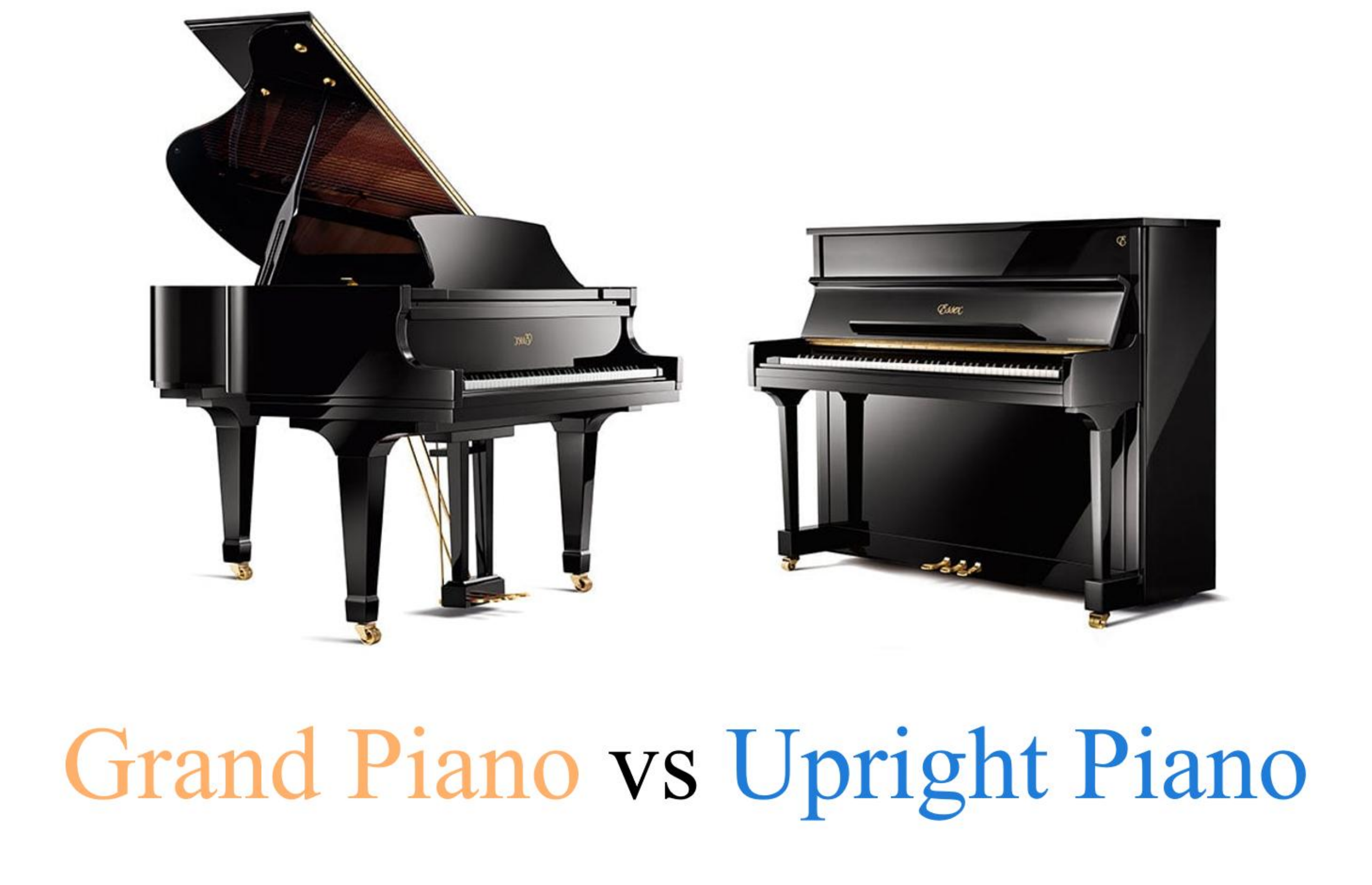}
    \end{minipage}
    \hfill
    \begin{minipage}[c]{0.68\linewidth}
        \centering
        \includegraphics[width=\linewidth]{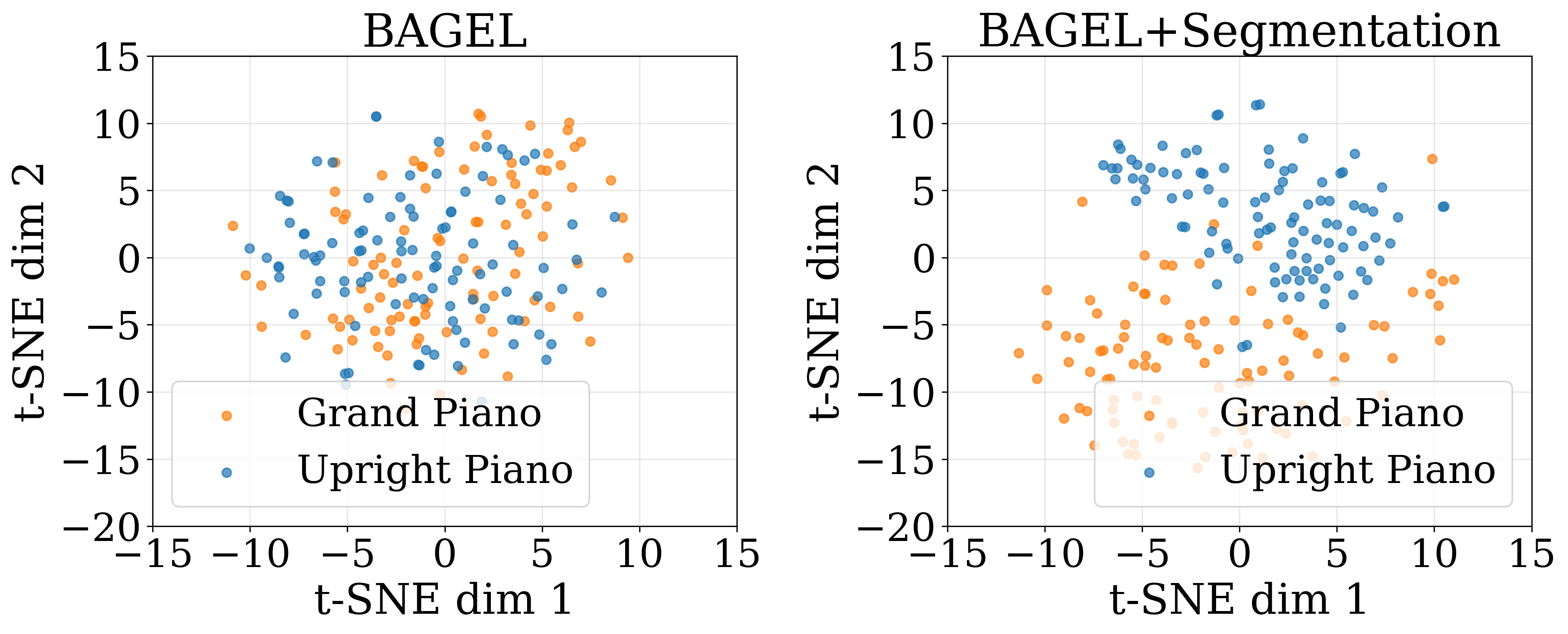}
    \end{minipage}
    \caption{\textbf{Feature space analysis on fine-grained classes.} 
    Left panels display the visually confusable categories \textcolor{orange}{Grand Piano} and \textcolor{eccvblue}{Upright Piano}. The corresponding tSNE visualizations on the right reveal that while the baseline BAGEL yields entangled feature spaces, our proposed BAGEL+Segmentation learns highly discriminative embeddings and achieves clear class separation.}
    \label{fig:tsne}
\end{figure}

To further elucidate the impact of the SGT, we employ BAGEL as a representative architecture to investigate specific representational shifts at both the feature and attention levels. Our analysis examines the model's internal dynamics across three dimensions encompassing the feature space structure of the visual encoder, the cross-modal attention patterns within the understanding module, and the attention distribution during generation.

\begin{figure}[t]
    \centering
    \begin{subfigure}[t]{0.58\textwidth}
        \centering
        \includegraphics[width=\textwidth]{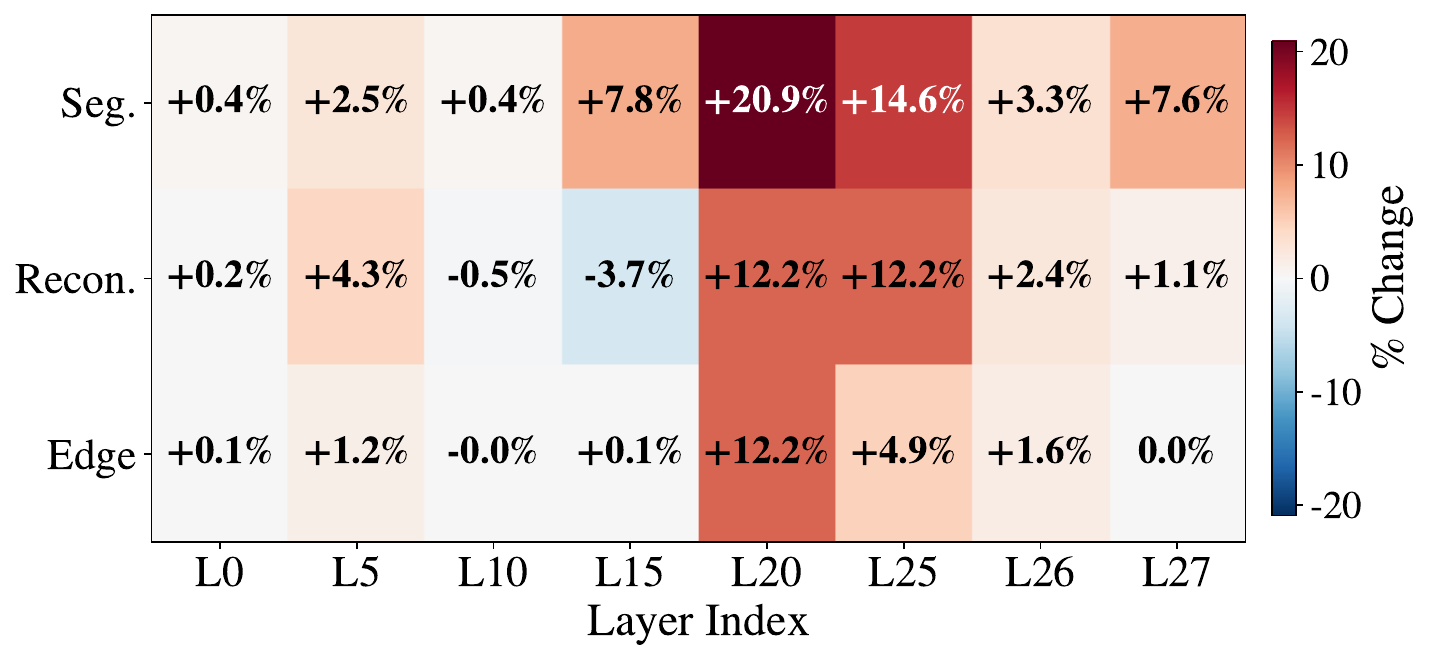}
        \caption{Vision-language attention allocation.}
        \label{fig:vl_layer_attention}
    \end{subfigure}
    \hfill
    \begin{subfigure}[t]{0.38\textwidth}
        \centering
        \includegraphics[width=\textwidth]{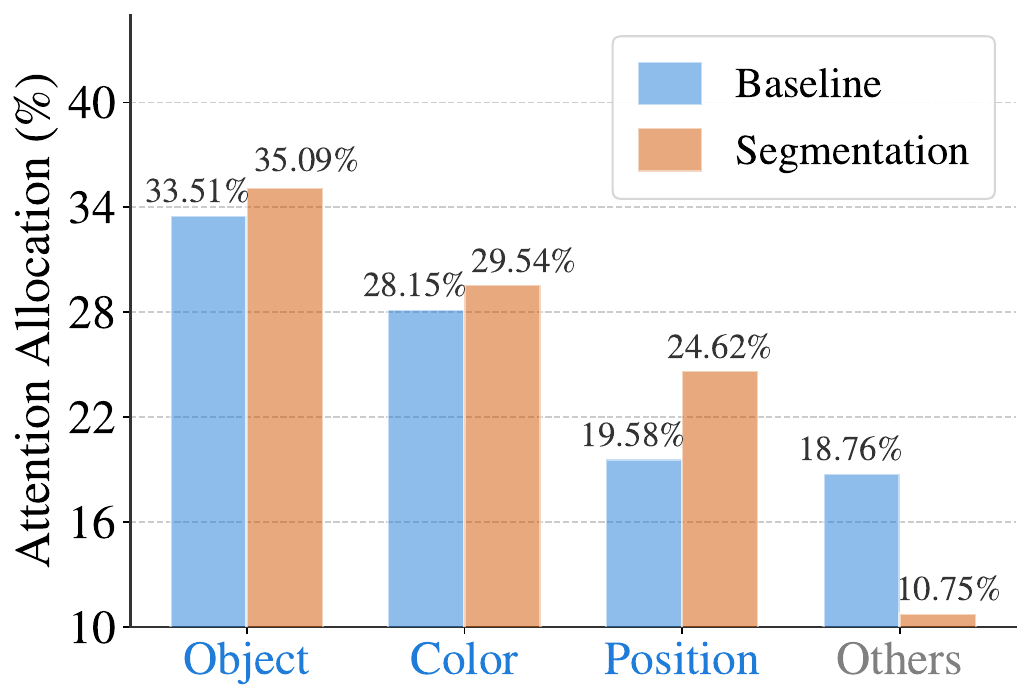}
        \caption{Key words attention allocation.}
        \label{fig:ug_attention_allocation}
    \end{subfigure}
    \caption{\textbf{Analysis of attention patterns.} 
    (a) Layer-wise changes in attention to visual features for three proxy tasks relative to the BAGEL baseline, demonstrating a consistent increase in visual focus in deeper layers.
    (b) Attention distribution over text tokens. The segmentation objective effectively enhancing the focus on critical tokens (\textcolor{eccvblue}{Object, Color, Relation}).}
    \label{fig:attention_analysis}

\end{figure}

\vspace{2.5pt}
\noindent\textbf{Finding 1: SGT promotes feature linear separability.} We first visualize the visual embeddings $z_{vit}$ using t-SNE as shown in Fig.~\ref{fig:tsne}. The projections reveal that training with segmentation data enhances the linear separability of categories that are semantically similar yet structurally distinct, such as upright and grand pianos. In contrast to the baseline model which often yields diffuse clusters, the incorporation of segmentation supervision significantly improves both the intra-class compactness and inter-class separability of the visual representations.

\vspace{2.5pt}
\noindent\textbf{Finding 2: Mitigating linguistic over-reliance.} We examine the cross-modal attention dynamics within the understanding module, as shown in Fig.~\ref{fig:vl_layer_attention}. Specifically, we observe a higher concentration of attention on visual tokens within the deeper transformer layers compared to the baseline. This distribution indicates that the model anchors its reasoning process more firmly in visual evidence, effectively counteracting the over-reliance on linguistic priors that often leads to hallucination~\cite{bias:Peng2022Balanced,bias:zheng2025mllms}. Crucially, high-level segmentation tasks induce a more pronounced attention shift than low-level objectives. 

\vspace{2.5pt}
\noindent\textbf{Finding 3: Amplifying critical tokens and suppressing irrelevant cues.} We investigate the generative capability using prompts containing position and attribute constraints sampled from GenEval~\cite{bench:geneval}. We quantify the cross-attention weights allocated to critical tokens, specifically position, color, and object identity. As illustrated in Fig.~\ref{fig:ug_attention_allocation}, the integration of segmentation data amplifies the model's focus on these attribute-specifying tokens. This further demonstrates that SGT effectively narrows the representational gap within UMMs and compels the models to prioritize intrinsically meaningful features.

\section{Limitations}
While SGT effectively aligns understanding and generation for natural scenes, relying exclusively on segmentation data constrains performance on symbolically dense and knowledge-intensive tasks, as shown in~\ref{fig:gain_understanding}. This observation indicates that SGT functions best as a foundational alignment strategy rather than a standalone training solution. The paradigm successfully retains its symbolic proficiency when SGT is augmented with VQA data, as shown in Table~\ref{tab:sota_comparison}. Future research will explore a comprehensive post-training pipeline integrating the SGT alignment strategy with understanding data, generative targets, and reinforcement learning frameworks to achieve optimal cross-modal performance.
\section{Conclusion}

This work proposes a fine-tuning paradigm for UMMs to mitigate the optimization divergence between visual understanding and generation. Previous attempts leverage pixel space reconstruction to improve multimodal alignment but inadvertently introduce granular visual noise that ultimately leads to suboptimal performance. To overcome this limitation, we introduce Semantic Generative Tuning as a novel paradigm that shifts the alignment proxy from the pixel space to the semantic space. Mechanistic analyses reveal that this semantic integration fundamentally improves feature linear separability and optimizes attention allocation to directly mitigate representational misalignment. Extensive empirical evaluations across mainstream architectures demonstrate that SGT consistently yields significant improvements in both visual understanding accuracy and generative layout fidelity. The principles established by this paradigm highlight that aligning multimodal capabilities at the semantic level serves as a crucial foundation for developing cohesive and versatile UMMs.

\clearpage
\section{Appendix}
\begin{itemize}
    \item \textbf{Section~\ref{supple:data}:} Data Processing
    \item \textbf{Section~\ref{supple:method}:} Method Details
    \item \textbf{Section~\ref{supple:config}:} Training Configurations
    \item \textbf{Section~\ref{supple:infer}:} Inference Settings and Additional Results
    \item \textbf{Section~\ref{supple:analysis}:} Mechanistic Analysis Methods
\end{itemize}

\subsection{Data Preparation}
\label{supple:data}
\begin{figure}[t]
    \centering
    \includegraphics[width=\textwidth]{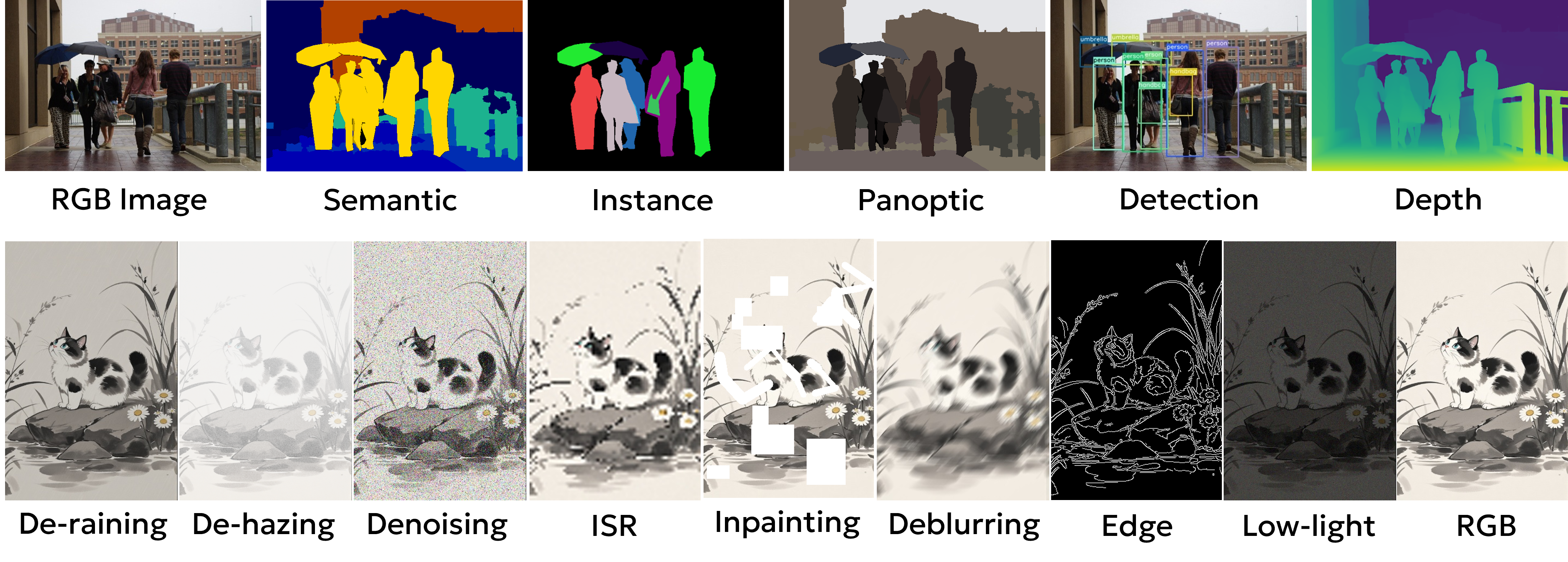}
    \caption{Illustration of various computer vision tasks. \textbf{Top row:} RGB Image, Semantic Segmentation, Instance Segmentation, Panoptic Segmentation, Object Detection, and Depth Estimation. \textbf{Bottom row:} (This figure serves solely illustrative purposes and does not originate from the MS COCO dataset.) De-raining, De-hazing, Denoising, Image Super-Resolution (ISR), Deblurring, Edge Detection, Low-light Enhancement, and RGB reference.}
    \label{fig:vision_tasks}
\end{figure}

This study systematically evaluates the impact of classic vision tasks on UMMs within the generative tuning framework. The evaluated tasks span from high-level segmentation and object detection to low-level edge detection and image super-resolution. The training set of MS COCO serves as the primary experimental basis to streamline data acquisition. Original ground truth annotations from this dataset provide the target labels for semantic segmentation, instance segmentation, panoptic segmentation and object detection. Training samples for the remaining visual tasks originate directly from the corresponding RGB images. Each individual task category consists of 20k sample pairs. Table~\ref{tab:vision_taxonomy} and Fig.~\ref{fig:vision_tasks} present the definitions and configurations of the various visual proxy tasks evaluated in our study.

\begin{table*}[h]
    \centering
    \caption{\textbf{Taxonomy of Computer Vision Tasks.} We summarize common vision tasks with their primary objectives and definitions.}
    \label{tab:vision_taxonomy}
    \resizebox{0.98\textwidth}{!}{%
        \begin{tabular}{l l}
            \toprule
            \textbf{Task} & \textbf{Primary Goal / Definition} \\
            \midrule
            Object Detection & Localize and classify objects with bounding boxes. \\
            Semantic Segmentation & Classify each pixel into a predefined category (no instance distinction). \\
            Instance Segmentation & Detect and segment each distinct object instance. \\
            Panoptic Segmentation & Unify semantic and instance segmentation (stuff + things). \\
            Edge Detection & Identify points in an image where brightness changes sharply. \\
            Depth Estimation & Predict the distance of each pixel relative to the camera. \\
            Image Denoising & Remove noise from images while preserving details. \\
            Image De-raining / De-hazing & Recover clear images from rain streaks or hazy conditions. \\
            Image Deblurring & Restore sharp images from motion or focal blur. \\
            Low-Light Enhancement & Improve visibility and contrast in dark/underexposed images. \\
            Image Super-Resolution (ISR) & Reconstruct high-resolution images from low-resolution inputs. \\
            Image Inpainting & Fill in missing or corrupted parts of an image. \\
            \bottomrule
        \end{tabular}%
    }
\end{table*}

\noindent\textbf{Segmentation \& Detection.}
Segmentation tasks rely directly on ground truth annotations from the MS COCO dataset for supervision. A colorization process transforms these original annotations into three-channel pseudo-color images to serve as the final target signals. In the case of object detection, bounding boxes along with their associated categorical labels are explicitly rendered onto the original images to establish the target representations.

\noindent\textbf{Depth Estimation.}
Ensuring the accuracy of depth annotations involves deploying both Depth Anything V2 and DepthPro to independently estimate depth maps for images from the MS COCO dataset. A least squares alignment then evaluates the consistency between these parallel estimations. Samples are discarded if the discrepancy between the two model outputs exceeds a predefined threshold of 0.4 following the alignment process. Random replacements from the broader dataset compensate for these discarded instances to maintain a constant overall training volume. The data overlap across all evaluated tasks exceeds 95\% to guarantee fair comparisons. The relative depth outputs from Depth Anything V2 are normalized and replicated three times along the channel dimension to serve as the final supervision targets.

\noindent\textbf{Edge Detection.}
The Canny edge detector extracts image edges to establish the ground truth for the edge detection task. The algorithm applies a lower threshold of 100 and an upper threshold of 200.

\noindent\textbf{Inpainting.}
The inpainting task reconstructs missing regions within RGB images. A masking procedure corrupts the original images using either random lines or solid blocks. The UMMs process these degraded images as input and learn to reconstruct the original RGB counterparts. The missing regions are randomly filled with either black or white pixels.

\noindent\textbf{Image Super-Resolusion.}
Image super-resolution tasks require the model to reconstruct a clear high-resolution image from a low-resolution input. We apply downsampling factors of 2, 4, 6 and 8 to generate the input training data for the UMMs. The generative tuning framework necessitates identical input and output resolutions. We therefore apply bilinear interpolation to the downsampled images to restore their spatial dimensions to match the target resolution before feeding them into the UMMs. This consecutive downsampling and upsampling procedure causes inevitable information loss and creates an information bottleneck that constitutes the primary challenge of super-resolution.

\noindent\textbf{Deblurring.}
Image deblurring aims to reconstruct a sharp image from a motion-blurred counterpart. A simulation algorithm artificially generates this degradation. The process applies random blur angles uniformly sampled from a full $360^{\circ}$ range and employs three distinct blur kernel sizes of 10, 20 and 30.

\noindent\textbf{Low-Light Image Enhancemen.}
Low-light image enhancement restores a high-quality image with proper exposure, sharp details and natural color distributions from an observation captured under insufficient illumination. A synthetic degradation pipeline simulates these conditions by randomly scaling the brightness of clean images with factors ranging from 0.1 to 0.5. The pipeline simultaneously introduces noise signals with intensities restricted between 0.01 and 0.04. The UMMs process these darkened and noisy images as input to reconstruct the original clean counterparts.

\noindent\textbf{Denoising.}
Image denoising aims to restore clean images from noisy observations. A stochastic degradation pipeline introduces synthetic noise into clean images to generate the corresponding degraded inputs. This process incorporates Gaussian noise, JPEG compression artifacts, salt-and-pepper noise and Poisson noise. The algorithm superimposes these degradation types with random probabilities and in a random order to synthesize the final corrupted images.

\begin{table}[t]
\centering
\caption{Summary of benchmarks for six core visual understanding capabilities.}
\label{tab:benchmarks}
\resizebox{\columnwidth}{!}{%
\begin{tabular}{lcccccc}
\toprule
\textbf{Capability} & \textbf{Vision-centric} & \textbf{Spatial Reasoning} & \textbf{Hallusion} & \textbf{General VQA} & \textbf{Chart/OCR} & \textbf{Knowledge/Math} \\
\midrule
\multirow{2}{*}{Benchmarks} 
    & CV-Bench~\cite{bench:CV-bench} & VSR~\cite{bench:VSR} & POPE~\cite{bench:pope} & MMMU~\cite{bench:mmmu} & OCRBench~\cite{bench:ocrbench} & MathVista~\cite{bench:mathvista} \\
    & MMVP~\cite{bench:MMVP} & SIBench~\cite{bench:sibench} & HallusionBench~\cite{bench:hallusion} & MMStar~\cite{bench:mmstar} & DocVQA~\cite{bench:docvqa} & ScienceQA~\cite{bench:scienceqa} \\
\bottomrule
\end{tabular}
}
\end{table}

\subsection{Detailed Results in Section~\ref{sec:medthod_3findings}.}
\label{supple:method}
\noindent\textbf{Benchmarks.} A comprehensive evaluation of visual understanding capabilities under generative tuning assesses six core competencies including vision centric perception, hallucination resistance, spatial reasoning, general visual question answering, document and chart comprehension, and mathematical and knowledge reasoning. The evaluation relies on diverse datasets comprising CV-Bench, MMVP, HallusionBench, POPE, SIBench-mini, MMMU-val, MMStar, DocVQA-val, ChartQA, Mathvista-mini and ScienceQA, as shown in Tabel~\ref{tab:benchmarks}. We consolidate specific low-level tasks within a single training phase by selecting between deraining and dehazing with equal probability. The training procedure similarly integrates image restoration objectives by randomly sampling among denoising, deblurring and low-light enhancement. 

\noindent\textbf{Detailed results of Fig.~\ref{fig:gain_understanding}.} Table~\ref{tab:method_results} presents the detailed experimental results of Fig.~\ref{fig:gain_understanding}. The segmentation task yields the most significant improvements for both BAGEL and OmniGen2 architectures. We therefore recommend adopting the segmentation task as the primary target for generative tuning and refer to this methodology as \textbf{S}emantic \textbf{G}enerative \textbf{T}uning. The term semantic in this context avoids specific categorical information and instead denotes a high-level representation of image content. We also observe that relying solely on generative tuning causes a slight degradation in table interpretation and knowledge reasoning capacities. We hypothesize that while generative tuning facilitates better alignment within the representation space of UMMs, it does not introduce supplementary logical reasoning skills or prior knowledge. The validation of the proposed method relies exclusively on generative tuning without incorporating any additional supervised fine-tuning data. Generative tuning on the segmentation task yields a 1\% overall performance gain even under this strictly constrained setting. This improvement represents the aggregated score across 12 distinct benchmarks to provide strong statistical evidence. These results demonstrate that SGT effectively enhances the perceptual understanding capabilities of UMMs.

\begin{table*}[t]
\centering
\caption{\textbf{Detailed quantitative results corresponding to Fig.~\ref{fig:gain_understanding} are provided.} The evaluated benchmarks from left to right consist of CV-Bench-2D, MMVP, VSR, SIBench-mini, POPE, HallusionBench, MMMU-val, MMStar, OCRBench, DocVQA-val, MathVista-mini, ScienceQA-val and the overall average score. The \textit{Mixed} row in the table reports the performance achieved by training the model on a combined dataset that integrates panoptic segmentation, image reconstruction and edge detection.}
\label{tab:method_results}
\resizebox{\textwidth}{!}{%
\begin{tabular}{l|cccccccccccc|c}
\toprule
\textbf{Model} & \textbf{CV-B} & \textbf{MMVP} & \textbf{VSR} & \textbf{SIB} & \textbf{POPE} & \textbf{Hall.} & \textbf{MMMU} & \textbf{MMS} & \textbf{OCR} & \textbf{Doc} & \textbf{Math} & \textbf{Sci} & \textbf{Overall} \\
\midrule
BAGEL & 73.2 & 83.0 & 80.4 & 49.0 & 85.7 & 68.3 & 46.8 & 67.5 & 81.0 & 94.0 & 72.2 & 95.9 & 74.8 \\
+Panoptic & 78.9 & 85.0 & 81.4 & 49.8 & 88.5 & 69.2 & 48.0 & 68.9 & 79.3 & 93.3 & 71.9 & 95.3 & 75.8 \\
+Instance & 78.2 & 84.7 & 81.2 & 49.7 & 86.9 & 70.1 & 49.4 & 68.9 & 78.9 & 93.4 & 70.5 & 95.2 & 75.6 \\
+Semantic & 78.4 & 85.3 & 80.6 & 50.1 & 87.3 & 69.8 & 47.4 & 67.9 & 81.0 & 93.3 & 69.8 & 95.2 & 75.5 \\
+Depth & 74.2 & 83.7 & 81.6 & 49.7 & 87.4 & 68.4 & 48.8 & 68.3 & 78.2 & 92.6 & 69.9 & 94.2 & 74.8 \\
+Inpainting & 75.9 & 85.3 & 81.1 & 50.0 & 86.9 & 70.2 & 48.0 & 68.2 & 78.6 & 93.4 & 68.5 & 96.0 & 75.2 \\
+Detection & 75.6 & 84.7 & 81.0 & 50.2 & 87.4 & 69.4 & 47.4 & 68.2 & 78.3 & 93.2 & 70.1 & 95.4 & 75.1 \\
+Reconstruction & 74.8 & 83.7 & 81.2 & 49.6 & 87.2 & 68.8 & 47.1 & 67.9 & 78.1 & 93.8 & 68.9 & 95.4 & 74.7 \\
+Edge & 73.8 & 83.3 & 79.7 & 49.6 & 87.4 & 67.0 & 47.4 & 66.9 & 79.9 & 94.2 & 69.5 & 94.5 & 74.4 \\
+Derain-Dehaze & 74.3 & 83.3 & 79.8 & 50.3 & 87.2 & 67.2 & 47.1 & 67.3 & 78.1 & 93.2 & 69.5 & 94.4 & 74.3 \\
+Denoise-Deblur-Enhance & 74.6 & 83.7 & 79.5 & 49.3 & 87.6 & 68.6 & 47.0 & 66.9 & 79.3 & 93.3 & 69.3 & 94.5 & 74.5 \\
+ISR & 74.6 & 83.7 & 79.5 & 49.3 & 87.6 & 68.6 & 47.0 & 66.9 & 79.3 & 93.3 & 69.3 & 94.5 & 74.5 \\
+Mixed & 74.8 & 83.3 & 81.6 & 50.0 & 88.0 & 68.4 & 48.8 & 68.3 & 79.1 & 92.6 & 70.1 & 94.2 & 74.9 \\
\midrule
OmniGen2 & 65.9 & 65.0 & 77.5 & 43.3 & 86.0 & 62.4 & 42.1 & 55.1 & 81.3 & 93.4 & 62.3 & 79.3 & 67.8 \\
+Panoptic & 68.3 & 68.3 & 79.1 & 44.7 & 86.1 & 66.2 & 46.0 & 56.8 & 78.6 & 92.3 & 62.8 & 78.2 & 69.0 \\
+Depth & 66.4 & 66.7 & 78.1 & 44.1 & 86.1 & 65.7 & 45.0 & 55.6 & 77.3 & 92.2 & 61.6 & 77.3 & 68.0 \\
+Recon & 66.8 & 66.0 & 77.8 & 43.7 & 86.3 & 65.3 & 45.2 & 56.1 & 77.9 & 92.4 & 61.6 & 77.9 & 68.1 \\
+Inpainting & 67.2 & 67.0 & 79.5 & 43.9 & 86.2 & 63.4 & 45.2 & 55.6 & 78.5 & 93.0 & 61.7 & 78.8 & 68.3 \\
+Edge & 66.8 & 66.7 & 78.1 & 43.6 & 85.9 & 61.2 & 44.9 & 55.6 & 79.1 & 92.7 & 64.5 & 75.0 & 67.9 \\

\bottomrule
\end{tabular}%
}
\vspace{-2mm}
\end{table*}

\begin{table*}[t]
  \centering
  \setlength{\tabcolsep}{6pt}
  \caption{\textbf{Detailed quantitative results corresponding to Fig.~\ref{fig:gain_generation} are provided.} All reported metrics represent the average values computed across twelve independent random seeds to ensure statistical objectivity.}
  \label{tab:details_of_geneval}
  \resizebox{\textwidth}{!}{%
  \begin{tabular}{lrrrrrrr}
    \toprule
    Method & position & colors & color\_attr & counting & single\_object & two\_object & Overall \\
    \midrule
    BAGEL            & 51.3 & 86.3 & 63.2 & 79.2 & 99.1 & 92.8 & 78.6 \\
    +Edge            & \textbf{57.3} & 85.1 & 66.1 & \textbf{86.2} & 98.8 & \textbf{95.0} & 81.4 \\
    +Reconstruction  & 57.1 & 88.3 & 68.2 & \textbf{86.2} & \textbf{99.6} & 91.9 & \textbf{81.9} \\
    +Depth           & \textbf{57.3} & 86.2 & \textbf{74.5} & 80.5 & 98.3 & 93.9 & 81.8 \\
    +Segmentation    & 56.6 & \textbf{89.4} & 69.4 & 83.7 & 98.4 & 93.9 & \textbf{81.9} \\
    \midrule
    OmniGen2         & 47.0 & 88.3 & 64.0 & 66.3 & \textbf{99.7} & 93.9 & 76.6 \\
    +Edge            & 53.2 & \textbf{90.4} & 63.2 & 68.7 & 98.7 & 93.9 & 78.0 \\
    +Reconstruction  & 55.5 & 86.2 & \textbf{76.3} & 66.3 & {99.6} & 88.9 & 78.9 \\
    +Depth           & \textbf{56.6} & 89.4 & 63.5 & \textbf{73.7} & {99.5} & \textbf{96.0} & \textbf{79.9} \\
    +Segmentation    & 53.3 & 86.2 & 70.2 & 71.2 & 98.7 & 94.0 & 78.9 \\
    \bottomrule
  \end{tabular}}
\end{table*}

\noindent\textbf{Mixed three-task training.} We investigate whether combining diverse vision tasks yields greater improvements than applying a single task. The experimental setup integrates data from panoptic segmentation, image reconstruction and edge detection. The total sample capacity remains at 20,000 instances distributed equally among the three categories. Table~\ref{tab:method_results} demonstrates that combining these three data types produces smaller performance gains compared to utilizing segmentation data exclusively under identical data volume constraints. These comparisons across individual and mixed tasks indicate that semantic perception constitutes the most critical factor for the comprehension capabilities of UMMs.

\noindent\textbf{Detailed results of Fig.~\ref{fig:gain_generation}.} We provide detailed validation results of Fig.~\ref{fig:gain_generation}. We conduct twelve independent random sampling iterations across all evaluated methods. Table~\ref{tab:details_of_geneval} report the averaged results from twelve random seeds to depict the underlying performance trends objectively.

\noindent\textbf{Semantics matters: semantic generative tuning bridges sparse textual and dense visual signals as an intermediate representation.} We further clarify the relationships and distinctions among semantic generative tuning, instruction tuning and image generation. Visual understanding tasks depend on cross entropy loss for text based supervision where text representations provide concentrated and sparse semantic information. Visual generation tasks derive their supervision signals from the entire image which constitutes a structured and dense signal. The separate reliance on these two representations fragments the training process and impedes the true synergistic potential between understanding and generation capabilities. \textcolor{eccvblue}{Semantic generative tuning provides a structured visual supervision signal and clusters visual features into meaningful semantic regions. This methodology serves as an intermediate representation between sparse textual signals and dense RGB signals to bridge the gap between the two modalities.} We recommend high-level visual perception tasks as proxy objectives to stimulate the synergistic capabilities of UMMs. Semantic generative tuning functions effectively as a proxy task to align the representation spaces of understanding and generation but it does not intrinsically introduce new knowledge, logical reasoning skills or improvements in raw image generation quality. We conclude that semantic generative tuning should not serve as an isolated supervision signal since integrating this method with both understanding and generation training data yields the maximum performance gains.

\subsection{Training Configurations}
\label{supple:config}
We fine-tune both OmniGen2 and BAGEL using the AdamW optimizer with $\beta_1=0.9$ and $\beta_2=0.95$, and a weight decay of 0.01. Both models follow a dual-module architecture comprising an understanding module for visual-semantic comprehension and a generation module for image synthesis. Specifically, OmniGen2 employs a 3B-parameter understanding module coupled with a 4B-parameter generation module (7B total), whereas BAGEL adopts a more heavyweight design with 7B parameters allocated to each module (14B total). To accommodate the difference in model capacity, we adopt a learning rate of $4\times10^{-4}$ for OmniGen2 and a lower rate of $1\times10^{-4}$ for BAGEL, with warmup periods of 300 and 1,000 steps respectively. OmniGen2 is trained for 2,500 steps over approximately 4 hours, while BAGEL requires 10,000 steps over 18 hours. Both models are trained with a global batch size of 60. Detailed configurations are summarized in Table~\ref{tab:training_config}.

\begin{table}[t]
\centering
\caption{Training configurations of OmniGen2 and BAGEL.}
\label{tab:training_config}
\resizebox{0.45\textwidth}{!}{%
\begin{tabular}{lcc}
\toprule
\textbf{Configuration} & \textbf{OmniGen2} & \textbf{BAGEL} \\
\midrule
Parameters & 3B + 4B & 7B + 7B \\
Optimizer & AdamW & AdamW \\
Learning Rate & $4\times10^{-4}$ & $1\times10^{-4}$ \\
$\beta$ & (0.9, 0.95) & (0.9, 0.95) \\
Weight Decay & 0.01 & 0.01 \\
Warmup Steps & 300 & 1000 \\
Training Steps & 2500 & 10000 \\
Global Batch Size & 60 & 60 \\
Training Time & 4 hours & 18 hours \\
\bottomrule
\end{tabular}}
\end{table}


\begin{table}[t]
\centering
\caption{More results from mixed SFT and SGT training. These results confirm that the integration of generative tuning does not induce performance degradation in knowledge-intensive or text-recognition tasks.}
\label{tab:more_results}
\resizebox{0.6\textwidth}{!}{%
\begin{tabular}{lccccc}
\toprule
Model & ScienceQA & OCRBench & DocVQA & SEED & DPGBench \\
\midrule
BAGEL      & 95.9 & \textbf{81.0} & 94.0 & 77.3 & 84.0 \\
SFT      & 95.9 & 80.8 & \textbf{94.0} & 78.0 & 82.7 \\
SFT+SGT  & \textbf{95.9} & \textbf{81.0} & 94.0 & \textbf{79.5} & \textbf{84.0} \\
\bottomrule
\end{tabular}}
\end{table}

\subsection{Inference}
\label{supple:infer}
\begin{figure}[t]
  \centering
  \includegraphics[width=\linewidth]{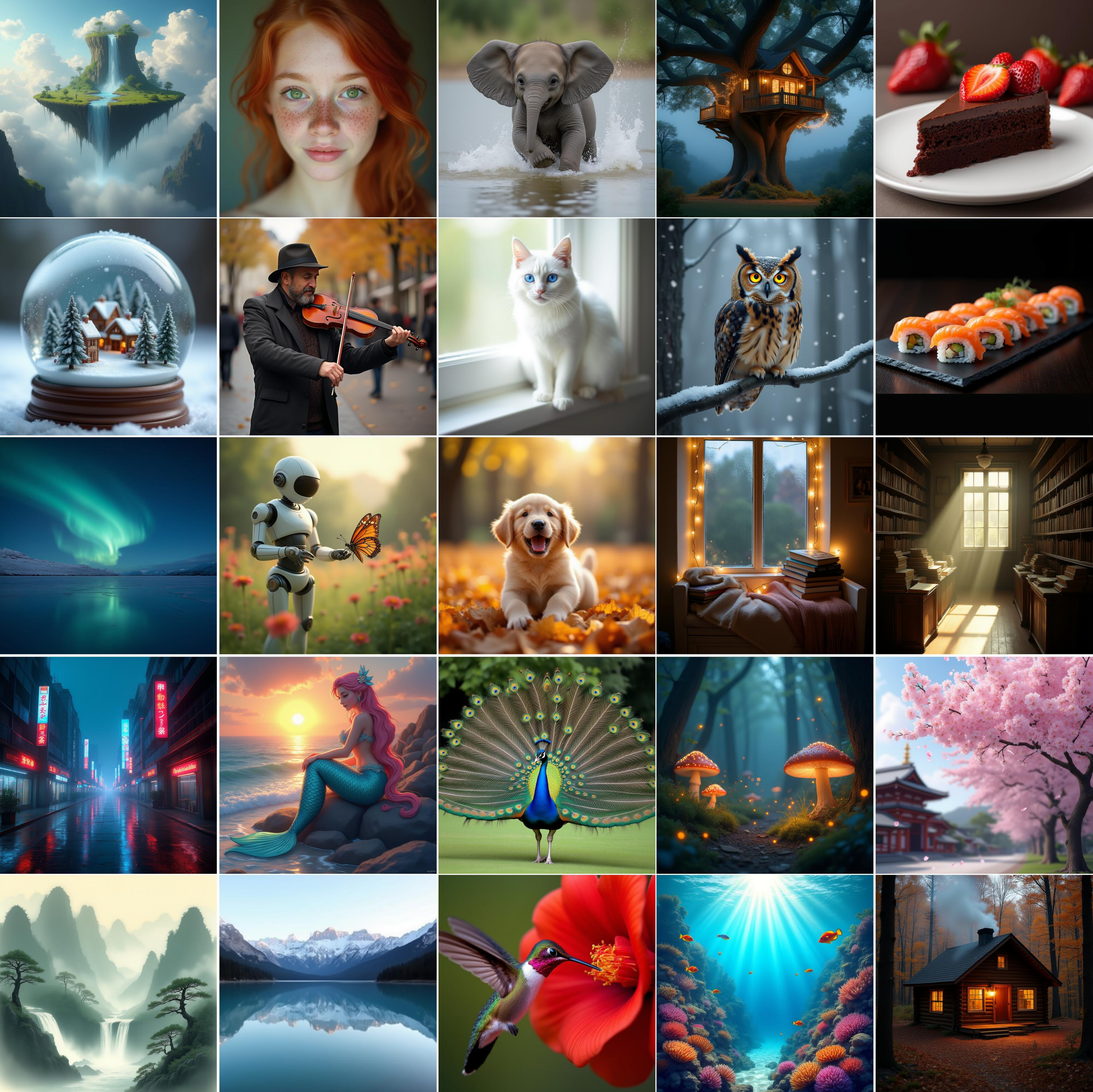} 
  \caption{Visualization of images generated by SGT, demonstrating high-quality and diverse generations across a wide range of prompts and scenes.}
  \label{fig:sgt-visualization}
\end{figure}

\noindent\textbf{Inference details.} The inference stage strictly follows the mechanism of the original model. Eq.~\ref{eq:umms} dictates that visual understanding tasks employ $f_{\theta}(x, [z_{vit}])$ to yield an output $y \in \mathcal{T}$. The framework processes text to image generation using $f_{\theta}(x, [z_{noise}])$ to produce $y \in \mathcal{I}$. For the BAGEL architecture, visual editing operations require $f_{\theta}(x, [z_{vit}, z_{vae}, z_{noise}])$ to generate the modified output $y \in \mathcal{I}$. Conversely, for OmniGen2, we adhere to the original official configuration for image editing tasks, utilizing $f_{\theta}(x, [z_{vae}, z_{noise}])$. Fig.~\ref{fig:sgt-visualization} displays samples produced by semantic generative tuning to illustrate the high fidelity of the synthesized images.

\noindent\textbf{Supplementary results.} Table~\ref{tab:more_results} provides supplementary evaluations that extend the results presented in Table~\ref{tab:sota_comparison} and Table~\ref{tab:main_results_ablation}. The experimental data indicates that the joint training of SFT and SGT datasets maintains the performance of the model in knowledge-intensive and OCR tasks compared to training solely on SFT data. Furthermore, assessments on the DPGBench benchmark reveal that the framework achieves consistent performance without significant improvement or degradation. DPGBench involves more extensive textual instructions compared to GenEval. The lack of substantial improvement on this benchmark suggests that the SGT framework does not inherently facilitate complex instruction parsing capabilities. Further enhancement of these editing proficiencies likely requires the integration of specialized generative datasets that are specifically curated for high-complexity instruction following.

\subsection{Mechanism Analysis}
\label{supple:analysis}
\noindent\textbf{tSNE.} We take BAGEL as a representative example to analyze the features extracted from its semantic vision encoder. Since BAGEL employs SigLIP2 as its vision encoder, which does not utilize a class token during training, we flatten all visual tokens into a single feature vector for each image. To enable effective visualization, we first apply Principal Component Analysis (PCA) to reduce the feature dimensionality to 50, followed by t-SNE projection onto a 2D plane for visualization. For t-SNE, we adopt the default perplexity value of 30. The complete pipeline is summarized in Fig.~\ref{fig:tsne_pseudocode}.

\begin{figure}[t]
\centering
\begin{tcolorbox}[colback=gray!5, colframe=gray!50, title=Pseudocode]
\begin{lstlisting}[style=mystyle-tsne]
# Feature Extraction
for image in dataset:
    feat = Flatten(model.vit(image))
    features.append(feat)
save(features)

# Dimensionality Reduction  
features_50d = PCA(features, 50)
features_2d  = tSNE(features_50d)

# Visualization
plot(features_2d, color=class_label)
\end{lstlisting}
\end{tcolorbox}
\caption{Pseudocode for t-SNE visualization pipeline.}
\label{fig:tsne_pseudocode}
\end{figure}

\begin{figure}[h]
    \centering
    \includegraphics[width=0.7\linewidth]{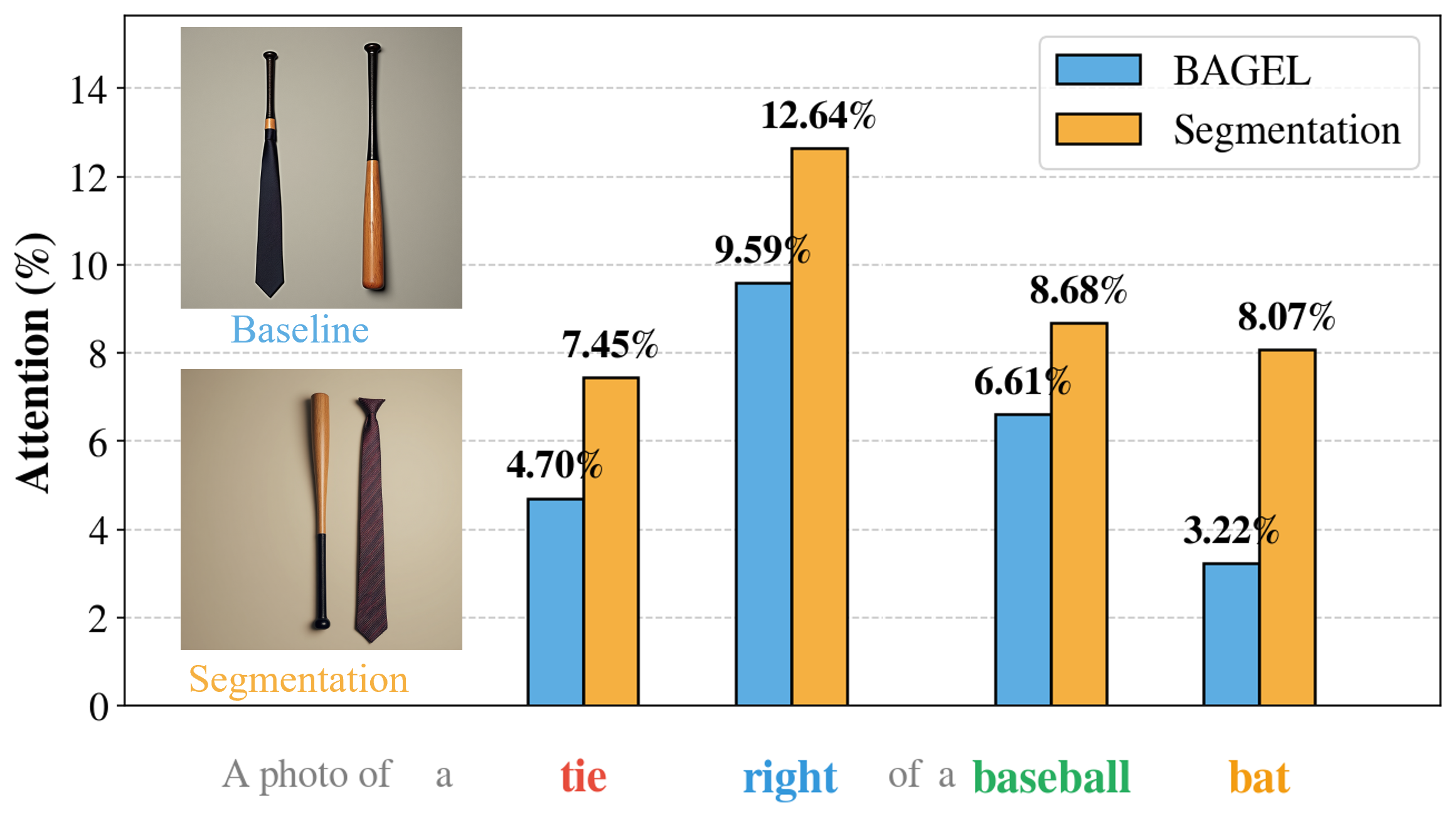}
    \caption{
        \textbf{Token-level attention distribution during image generation.}
        We visualize the attention weights allocated to each token in the prompt 
        ``\textit{A photo of a \textcolor{red}{tie} \textcolor{blue}{right} of a 
        \textcolor{green}{baseball} \textcolor{orange}{bat}}'' for both the baseline 
        BAGEL model and our segmentation-enhanced variant. 
        The segmentation guidance consistently amplifies attention to semantically 
        salient tokens (\textcolor{red}{tie}: $4.70\%{\to}7.45\%$, 
        \textcolor{blue}{right}: $9.59\%{\to}12.64\%$), leading to improved 
        spatial reasoning and object placement as shown in the generated samples (left).
    }
    \label{fig:ug_attention_distribution}
\end{figure}

\noindent\textbf{Key words attention allocation.}
To analyze the attention distribution over keywords during generation, we curate a diagnostic set of 20 prompts containing explicit spatial and color attributes. In BAGEL's flow-based generation, the noisy latent tokens serve as queries while the text prompt provides keys and values. We categorize the prompt tokens into four semantic groups: \textit{object} (e.g., nouns denoting entities), \textit{position} (e.g., spatial descriptors), \textit{color} (e.g., chromatic attributes), and \textit{others} (e.g., ``a'', ``the'', ``of''). For each category, we compute its relative attention weight as the proportion of total attention mass. Since early denoising steps are known to establish global semantic structure, we report the average attention distribution over the first three timesteps to capture the critical semantic binding phase. The complete procedure is summarized in Fig.~\ref{fig:keyword_attention_pseudocode}. Fig.~\ref{fig:ug_attention_distribution} illustrates the variations following semantic generative tuning. A statistical analysis on a sampled subset yields the results presented in Fig.~\ref{fig:ug_attention_allocation}. These findings demonstrate that the model concentrates more effectively on keywords after undergoing semantic generative tuning.

\begin{figure}[t]
\centering
\begin{tcolorbox}[colback=gray!5, colframe=gray!50, title=Pseudocode]
\begin{lstlisting}[style=mystyle-ug]
# Step 1: Keyword Extraction
keywords = extract_keywords(prompt)
token_ids = tokenizer.encode(prompt)
keyword_indices = {}  # keyword -> token positions
for kw in keywords:
    keyword_indices[kw] = find_token_positions(kw, token_ids)

# Step 2: Attention Map Computation (GQA)
def compute_attention(hidden_states, k_cache, q_proj):
    Q = q_proj(hidden_states)  # [seq_len, num_heads, head_dim]
    K = repeat_kv(k_cache, num_heads // num_kv_heads)
    scores = Q @ K.T / sqrt(head_dim)
    attn_map = softmax(scores, dim=-1)  # [num_heads, q_len, kv_len]
    return attn_map

# Step 3: Keyword Attention Analysis
def analyze_keyword_attention(attn_map, latent_indices, keyword_indices):
    # Extract latent-to-KV attention
    latent_attn = attn_map[:, latent_indices, :]
    
    attn_mean = mean(latent_attn, dim=[0, 1]) # [kv_len]
    
    # Aggregate attention per keyword
    keyword_attention = {}
    total_attn = sum(attn_mean)
    for kw, indices in keyword_indices.items():
        kw_attn = sum(attn_mean[indices])
        # percentage
        keyword_attention[kw] = kw_attn / total_attn * 100  
        
    
    return keyword_attention

# Step 4: Track During Generation
for t in timesteps:
    for layer in selected_layers:
        attn_map = compute_attention(latent_hidden, kv_cache[layer])
        kw_attn = analyze_keyword_attention(attn_map, latent_idx, kw_idx)
        log(timestep=t, layer=layer, keyword_attention=kw_attn)
\end{lstlisting}
\end{tcolorbox}
\caption{Pseudocode for keyword-level attention analysis during image generation. We extract keywords from the prompt, compute GQA attention maps at selected timesteps and layers, and aggregate attention scores for each keyword to quantify its influence on the generated image.}
\label{fig:keyword_attention_pseudocode}
\end{figure}

\clearpage

\bibliographystyle{splncs04}
\bibliography{main}
\end{document}